\documentclass[10pt,twocolumn,letterpaper]{article}

\usepackage{cvpr}              

\usepackage{graphicx}
\usepackage{amsmath}
\usepackage{amssymb}
\usepackage{booktabs}
\usepackage[table]{xcolor}
\usepackage{lipsum}
\usepackage{rotating}
\usepackage{mathtools}
\usepackage{bbm}
\usepackage{multirow}
\usepackage{wrapfig}
\usepackage{tabularx}
\usepackage[pagebackref,breaklinks,colorlinks]{hyperref}

\usepackage[capitalize]{cleveref}
\crefname{section}{Sec.}{Secs.}
\Crefname{section}{Section}{Sections}
\Crefname{table}{Table}{Tables}
\crefname{table}{Tab.}{Tabs.}

\definecolor{Gray}{gray}{0.95}

\def\cvprPaperID{5966} 
\def\confName{CVPR}
\def\confYear{2022}

\begin{document}

\title{Parameter-free Online Test-time Adaptation}
\author{
Malik Boudiaf\\
\'ETS Montreal \thanks{Work done as part of a research internship at FiveAI.
 Corresponding author: malik.boudiaf.1@etsmtl.net} \\
\and
Romain Mueller\\
FiveAI\\
\and
Ismail Ben Ayed\\
\'ETS Montreal\\
\and
Luca Bertinetto\\
FiveAI
}
\maketitle

\newcommand{\luca}[1]{\textcolor{orange}{L: #1}}
\newcommand{\malik}[1]{\textcolor{blue}{Malik: #1}}
\newcommand{\drafty}[1]{\textcolor{gray}{#1}}
\newcommand{\todo}[1]{\textcolor{red}{#1}}
\newcommand{\orange}[2]{\textcolor<#1>{orange}{#2}}

\definecolor{lightgray}{gray}{0.55}
\newcommand{\result}[2]{ #1 \color{lightgray}{\scriptsize{$\pm{#2}$}}}
\newcommand{\magicpar}[1]{\smallskip\noindent {\textbf{#1}}\enskip}
\newenvironment{tight_enum}{
		\begin{enumerate}
        \setlength{\itemsep}{0pt}
        \setlength{\parskip}{0pt}
        \setlength{\parsep}{0pt}
}{\end{enumerate}}

\newenvironment{tight_it}{
\begin{itemize}
        \setlength{\itemsep}{0pt}
        \setlength{\parskip}{0pt}
        \setlength{\parsep}{0pt}
}{\end{itemize}}

\def\@onedot{\ifx\@let@token.\else.\null\fi\xspace}
\def\iid{i.i.d\onedot} 
\def\wrt{w.r.t\onedot} 
\newcommand{\LAME}{\textsc{LAME}\xspace}

\newcommand{\xbf}{\mathbf{x}}
\newcommand{\zbf}{\mathbf{z}}
\newcommand{\pbf}{\mathbf{p}}
\newcommand{\qbf}{\mathbf{q}}
\newcommand{\lbf}{\boldsymbol{\lambda}}
\newcommand{\zbft}{\tilde{\mathbf{z}}}
\newcommand{\XBF}{\mathbf{X}}
\newcommand{\ZBF}{\mathbf{Z}}
\newcommand{\ZBFT}{\tilde{\mathbf{Z}}}
\newcommand{\KL}{\textsc{KL}}
\newcommand{\zt}{\tilde{\mathbf{z}}}
\newcommand{\thetabf}{\boldsymbol{\theta}}
\newcommand{\norm}[1]{\left\lVert#1\right\rVert}
\newcommand{\ceq}{\stackrel{\mathclap{\normalfont\mbox{c}}}{=}}
\newcommand{\cleq}{\stackrel{\mathclap{\normalfont\mbox{c}}}{\leq}}
\newcommand{\ones}{\mathbbm{1}}
\let\oldemptyset\emptyset
\let\emptyset\varnothing

\begin{abstract}
Training state-of-the-art vision models has become prohibitively expensive for researchers and practitioners.
For the sake of accessibility and resource reuse, it is important to focus on adapting these models to a variety of downstream scenarios.
An interesting and practical paradigm is online test-time adaptation, according to which training data is inaccessible, no labelled data from the test distribution is available, and adaptation can only happen at test time and on a handful of samples.
In this paper, we investigate how test-time adaptation methods fare for a number of pre-trained models on a variety of real-world scenarios, significantly extending the way they have been originally evaluated.
We show that they perform well only in narrowly-defined experimental setups and sometimes fail catastrophically when their hyperparameters are not selected for the same scenario in which they are being tested.
Motivated by the inherent uncertainty around the conditions that will ultimately be encountered at test time, we propose a particularly ``conservative'' approach, which addresses the problem with a Laplacian Adjusted  Maximum-likelihood Estimation (LAME) objective.
By adapting the model's output (not its parameters), and solving our objective with an efficient concave-convex procedure, our approach exhibits a much higher average accuracy across scenarios than existing methods, while being notably faster and have a much lower memory footprint. The code is available at \url{https://github.com/fiveai/LAME}.
\end{abstract}

\section{Introduction}
\label{sec:intro}
In recent years, training state-of-the-art models has become a massive computational endeavor for many machine learning problems (\eg~\cite{gpt3_brown2020language,vit_dosovitskiy2020image,clip_radford2021learning}).
For instance, it has been estimated that each training of GPT-3~\cite{gpt3_brown2020language} produces an equivalent of 552 tons of CO2, which is approximately the amount emitted in six flights from New York to San Francisco~\cite{patterson2021carbon}.
As implied in the whitepaper on ``foundation models''~\cite{foundation_models_bommasani2021opportunities}, we should expect that more and more efforts will be dedicated to the design of procedures that allow for the efficient adaptation of pre-trained large models under a variety of circumstances.
In other words, these models will be ``trained once'' on a vast dataset and then adapted at test time to newly-encountered scenarios.
Besides being important for resource reuse, being able to abstract the \emph{pre-training stage} away from the \emph{adaptation} is paramount in privacy-focused applications, and in any other situation in which preventing access to the training data is desirable.
Towards this goal, it is important that, from the point of view of the adaptation system, there is neither access to the training data nor the training procedure of the model to adapt.
With this context in mind, we are particularly interested in designing adaptation methods ready to be used in realistic scenarios, and that are suitable for a variety of models.

One aspect that many real-world applications have in common is the need to perform adaptation \emph{online}, and with a limited amount of data.
That is, we should be able to perform adaptation while the data is being received.
Take for instance the vision model with which an autonomous vehicle or a drone may be equipped.
At test-time, it will ingest a video stream of highly-correlated data (non-\iid), which could be used for adaptation.
We would like to be confident that leveraging this information will be useful, and not destructive, no matter the type of domain shift that may exist between training and test data.
Such shifts could be, for instance, ``low-level'' (\eg the data stream is affected by snowy weather which has never been encountered during the California-sunlit training stage), or ``high-level'' (\eg the data include the particular Art Deco architecture of Miami Beach's Historic District), or even a combination of both.
To summarize, we are interested in the design of test-time adaptation systems that 
1) are unsupervised; 
2) can operate online and on potentially non-i.i.d.~data;
3) assume no knowledge of training data or training procedure; and
4) are not tailored to a certain model, so that the progress made by the community can be directly harnessed.

This problem specification falls under the \emph{fully test-time adaptation} paradigm studied in a handful of recent works~\cite{tent,ada_bn,shot,azimi2022self}, where simple techniques like test-time learning of batch normalization's scale and bias parameters~\cite{tent} have proven to be very effective in some scenarios, like the one represented by low-level corruptions~\cite{imagenet_c}.
In our experimental results, we observe that existing methods~\cite{tent,shot,ada_bn,pseudo_label} have to be used with great care in uncertain yet realistic situations because of their sensitivity to variables such as the model to adapt or the type of domain shift.
As a matter of fact, we show that, when selecting their hyperparameters to maximize the average accuracy over a number of scenarios, existing methods do not outperform a non-adaptive baseline. For them to perform well, hyperparameters need to be adjusted in a scenario-specific fashion.
However, this is clearly not an option when the test-time conditions are unknown in advance.

These findings suggest that, while being agnostic to both training and testing circumstances is important, it is wise to approach the problem of test-time adaptation prudently.
Instead of adapting the parameters of a pre-trained model, we only adapt its \emph{output} by finding the latent assignments that optimize a manifold-regularized likelihood of the data. The manifold-smoothness assumption has been successful in a wide range of other problems, including graph clustering \cite{shiMalik2000,shaham2018spectralnet,Tang2019KernelCK}, semi-supervised learning \cite{belkin2006manifold,Chapelle2010,IscenTAC19}, and few-shot learning~\cite{ziko2020laplacian}, as it enforces desirable and general properties on the solutions.
Specifically,  
we embed Laplacian regularization as a corrective term, and derive an efficient concave-convex procedure for optimizing our overall objective, with guaranteed convergence. When aggregating over different conditions, this simple and ``conservative'' strategy significantly improves both over the non-adaptive baseline and existing test-time adaptation methods in an extensive set of experiments covering 7 datasets, 19 shifts, 3 training strategies and 5 network architectures. Moreover, by virtue of not performing model adaptation but only output correction, it reduces \textit{by half} both the total inference time and the memory footprint compared to existing methods.


\section{Related work}
\label{sec:related_work}
In general, domain adaptation aims at relaxing the assumption that ``train and test distributions should match'', which is at the foundation of most machine learning algorithms.
Since real-world applications rarely reflect the textbook assumption, this relaxation has generated a lot of interest and motivated a large corpus of work.
Doing this topic justice would take several surveys (\eg~\cite{wilson2020survey,wang2021generalizing,csurka2017domain,patel2015visual}), and it is unfeasible given this paper format.
Instead, in this section we aim at describing the overall problem setups that are more closely relevant to ours.

The applicability of early works in domain adaptation was limited, in that methods required access to the target domain~\cite{patel2015visual} during training.
\textbf{Unsupervised domain adaptation}~\cite{wilson2020survey} makes the scenario slightly more realistic by not requiring labels from the target domain.
Two common general strategies are, for instance, explicitly learning domain-invariant feature representations by minimizing some measure of divergence between source and target distributions (\eg~\cite{long2015learning,sun2016deep,kang2019contrastive});
or embedding a ``domain discriminator'' component in the network and then penalizing its success in the loss (\eg~\cite{ganin2015unsupervised,purushotham2016variational}).
Still, the necessity of having access, during training, to \textit{both} source and target domains limits the usability of this class of methods.

\textbf{Domain generalization} (DG) foregoes the need to access the target distribution by learning a model from multiple domains, with the intent of generalizing to unseen ones~\cite{wang2021generalizing}.
Popular strategies to address this problem include:
increasing the diversity of training data via either augmentations (\eg~\cite{tobin2017domain,prakash2019structured}), adversarial learning (\eg~\cite{volpigeneralizing,zhou2020deep}), or generative models (\eg~\cite{rahman2019multi,somavarapu2020frustratingly}); learning domain-invariant representations~\cite{ben2007analysis}, and decoupling the domain-specific and domain-independent components (\eg~\cite{khosla2012undoing,niu2015multi,ilse2020diva}).
Notably, the recent work of Gulrajani \& Lopez-Paz~\cite{gulrajani2021search} showed on a large testbed that learning a vanilla classifier on a pool of datasets outperformed all modern techniques, thus sending a strong message on the importance of a carefully designed experimental protocol.

Despite the shared goal of generalizing across domains and the constraint of not having access to the target distributions in advance, one fundamental difference of DG with the setup we consider is the lack of test-time adaptability.
Instead, methods falling under the \textbf{source-free domain adaptation} paradigm~\cite{chidlovskii2016domain} require no access to the training data \emph{during the process of adaptation}.
Liang~\etal~\cite{liang2019distant} assume only to have access to the source dataset's summary statistics, and relate the models fitting the source and target domains by surmising that class centroids are only moderately shifted between the two datasets.
Before adaptation, Kundu~\etal~\cite{kundu2020towards,kundu2020universal} consider a first ``vendor-side'' phase, during which the target domain is not known and a model is trained on an augmented training dataset aiming at mimicking possible domain shifts and category gaps that will be encountered downstream.
Li~\etal~\cite{li2020model} propose the Collaborative Class Conditional GAN, which integrates the output of a prediction model into the loss of the generator to produce new samples in the style of the target domain, which are in turn used to adapt the model via backpropagation.
In \textit{Test-time Training}~\cite{sun2020test}, Sun~\etal perform test-time adaptation via self-supervision by jointly optimizing two branches (one supervised and one self-supervised) during training.

While being vastly more practical than the ones addressing vanilla domain adaptation, the methods listed above are still quite limited in that they typically have an ad-hoc training procedure.
As mentioned in Section~\ref{sec:intro}, we would like to facilitate model reuse, so that the progress made by the community in architecture design~\cite{vit_dosovitskiy2020image}, self-supervised learning~\cite{chen2020simple} or multi-modal learning~\cite{clip_radford2021learning} can be directly exploited.
Our setup is mostly similar to what has been referred to in the TENT paper~\cite{tent} as the \textbf{\textit{fully} test-time adaptation} scenario.
In this case, the intent is to perform unsupervised test-time adaptation while \textit{``not restricting or altering model training''}~\cite{tent}.
In TENT, this is achieved with a simple entropy minimization loss, which informs the optimization of scale and bias parameters of batch normalization layers.
As for batch normalization layers' statistics, they are re-estimated on the test data, similarly to what is done in adaptive batchnorm (AdaBN) methods~\cite{ada_bn,schneider2020improving,nado2020evaluating,burns2021limitations}, which have shown strong performance on the perturbations of ImageNet-C~\cite{imagenet_c}.
In similar spirit, Liang~\etal~\cite{shot} updates the parameters of the feature extractor of a given model by maximizing a mutual information objective (SHOT-IM).

Although we share many of the motivations presented in TENT and SHOT, we believe that our work differs under two main aspects.
First, given our model-independence desideratum, we explicitly study the extent to which our approach works across training strategies and architectures.
This analysis is missing in prior works: as we will see in Section~\ref{sec:experiments}, the type of model being adapted is a variable that strongly affects the effectiveness of both TENT and SHOT.
Second, for the sake of usability, we are particularly focused on \emph{online} adaptation, which leads us to also consider non-\iid scenarios as an important part of our evaluation.


\section{Problem Formulation}
\label{sec:problem_formulation}
\begin{figure*}[t!]
	\includegraphics[width=0.32\textwidth]{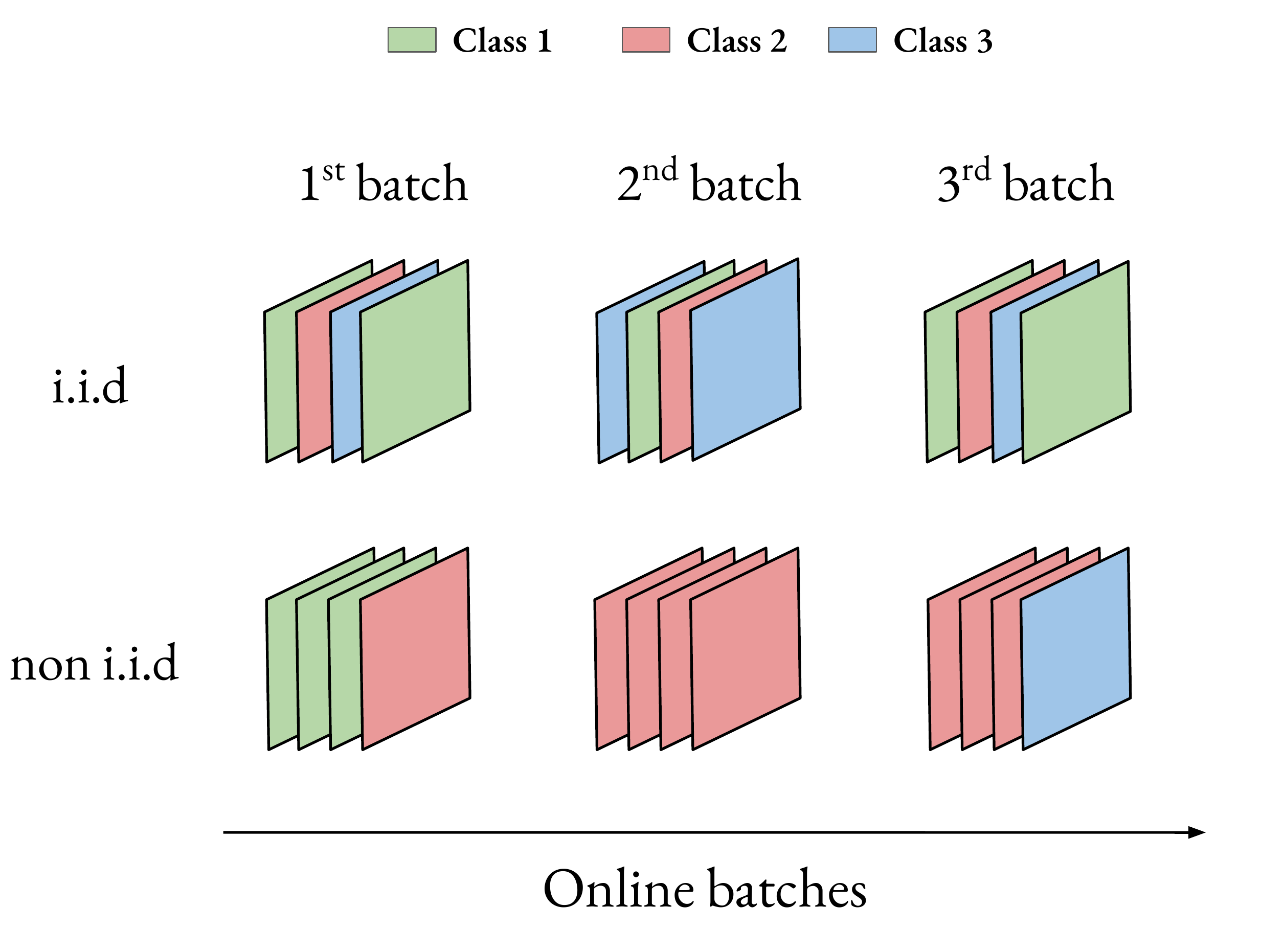}
	\includegraphics[width=0.32\textwidth]{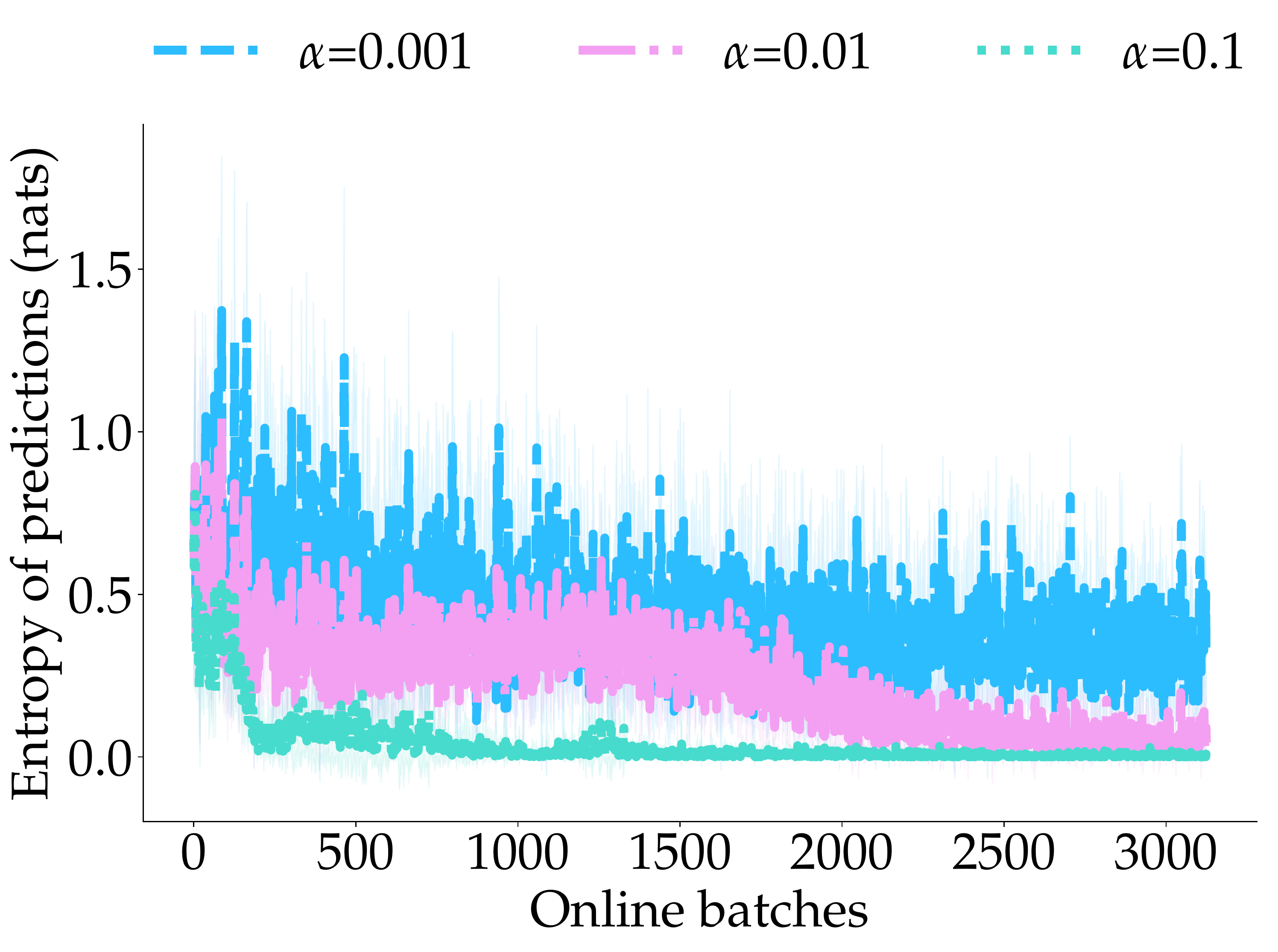}
	\includegraphics[width=0.32\textwidth]{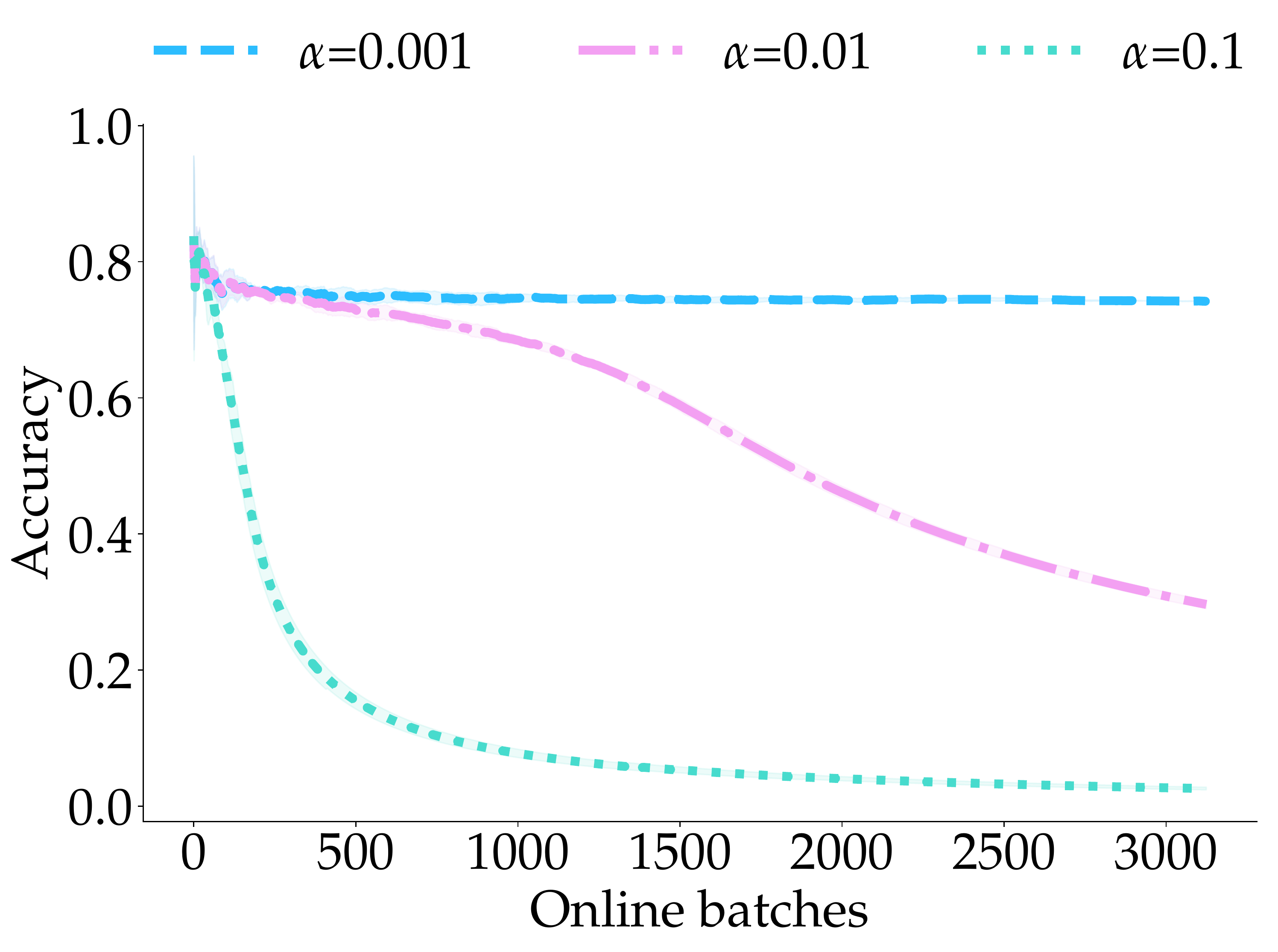}
	\caption{\textbf{Adaption through entropy minimization in a non-\iid scenario may \textit{silently} degenerate the model.} (Left) Non-\iid streams are generated by batching samples according to their class. (Middle) The conditional entropy of predictions is being minimized in an online fashion on such non-\iid streams. However, assessing whether the adaptation is being beneficial or detrimental solely from these curves is impractical in an unsupervised scenario. (Right) Rather, monitoring the online accuracy (which would require access to the labels) would reveal that the model is actually collapsing for two out of three learning rates considered.}
	\label{fig:non_iid}
\end{figure*}

\begin{figure}[t!]
    \centering
	\includegraphics[width=0.8\columnwidth]{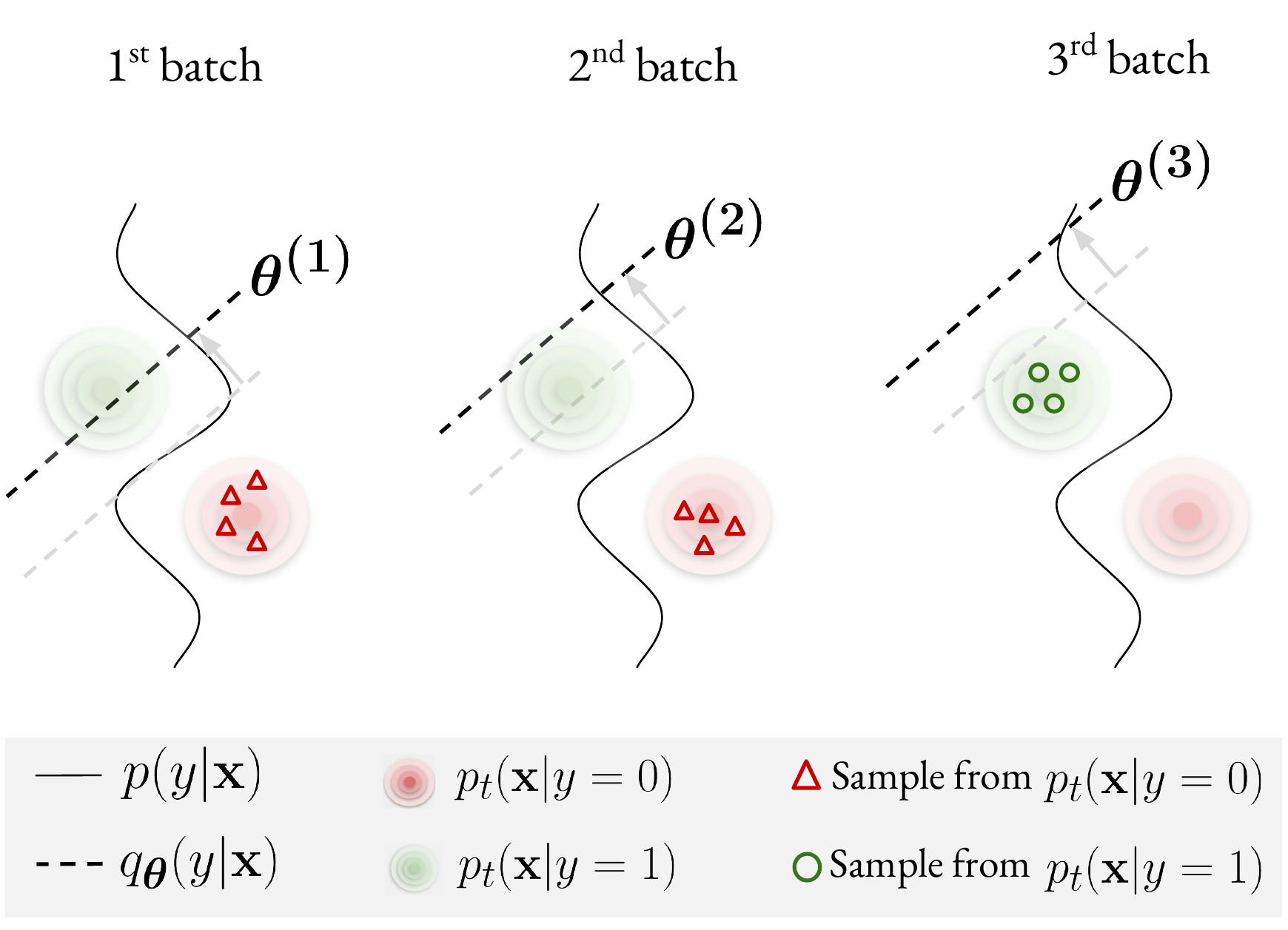}
	\caption{
	Minimizing the conditional entropy (as in TENT~\cite{tent}) encourages the model $q_{\thetabf}(y|x)$ to produce high confidence predictions.
	Geometrically, this corresponds to increasing the margin between the decision boundary and the samples from the current batch.
	In the low-diversity scenario depicted above, the $1^{st}$ and $2^{nd}$ batches only contain red samples.
	This causes the boundary to move away from red samples.
	When green samples are finally observed in the 3$^{rd}$ batch, the boundary has gone past the green cluster, so that samples are (wrongly) assigned to the red class.
	}
	\label{fig:non_iid_explanation}
\end{figure}


In (fully) test-time adaptation \cite{tent, shot} (TTA), we have access to a parametric model $q_{\thetabf}(y|\xbf)$ trained on an inaccessible labelled source dataset $\mathcal{D}_s = \{(\xbf, y) \sim p_s(\xbf,y) \}$, where $\xbf$ is an image and $y \in \mathcal{Y}$ its associated label from the set of source classes $\mathcal{Y}$. Additionally, we consider an unlabelled target dataset sampled from an arbitrary target distribution ${\mathcal{D}_t = \{\xbf \sim p_t(\xbf) \}}$.
We take the standard \textit{covariate shift} assumption \cite{storkey2009training} that $p_s(y|\xbf)=p_t(y|\xbf)$ and $p_s(\xbf) \neq p_t(\xbf)$, which implies that shifts can only happen if there exists some class $y$ such that $p_s(y)p_s(\xbf|y) \neq p_t(y)p_t(\xbf|y)$.
This leads us to consider two types of shift throughout this work: the \textit{prior shift}, in which $p_t(y)$ differs from $p_s(y)$, and the \textit{likelihood shift}, in which $p_t(\xbf | y)$ differs from $p_s(\xbf | y)$.

As the target distribution shifts from the source, the parametric model $q_{\thetabf}(y|\xbf)$ no longer necessarily well approximates the true, domain-invariant distribution $p(y|\xbf)$.
A toy illustration of this phenomenon can be found in Fig.~\ref{fig:non_iid_explanation}, where the linear classifier can only properly model the true sinusoidal distribution over a limited region of the input space.
Therefore, TTA methods aim at adapting $q_{\thetabf}(y|\xbf)$ to maximize its predictive performance on the target distribution.
In particular, we focus on the \textit{online} setting, where the classifier receives a potentially non-\iid stream of target samples, and must simultaneously adapt and predict.

Typical large-scale datasets contain up to tens of thousands of classes, and have been created with the purpose of covering a large portion of the concepts that may be of interest at test-time.
As such, they likely contain classes of a finer or equal (but not coarser) granularity than those required in specific TTA scenarios.
Therefore, to make our setting more practical, we relax the common assumption that source classes must coincide with the target ones.
Instead, we allow target classes to be superclasses, according to some pre-defined hierarchy. Authors from \cite{taori2020measuring} handle this by max-pooling the softmax predictions across associated subclasses, but we empirically found average-pooling to perform slightly better, and decided to proceed with this strategy. More details in Appendix.


\section{On the Risks of Network Adaptation}
\label{sec:network_adaptation}

In order to better approximate the underlying distribution $p(z|\textbf{x})$  at test-time, TTA methods usually propose to directly modify the parametric source model.
We group such methods under the term \textit{Network Adaptation Methods} (NAMs).
Specifically, such methods \cite{tent, shot} first partition the network into \textit{adaptable} weights $\thetabf^a$ and frozen weights $\thetabf^f$, and proceed by minimizing an unsupervised loss $\mathcal{L}(\xbf ; \thetabf^a \cup \thetabf^f), \ \xbf \sim p_t(\xbf)$ \wrt $\thetabf^a$.
TTA methods mostly differ based on their choices of partition $\{\thetabf^f, \thetabf^a\}$ and loss function $\mathcal{L}$.
For instance, TENT~\cite{tent} only adapts the scale and bias parameters $(\gamma, \beta)$ of the batch normalization (BN) layers through entropy minimization, while SHOT \cite{shot} adapts the convolutional filters of the model through mutual information maximization. 

While NAMs have the potential to 
substantially improve the performance of a model on the target samples, they also run the risk of dramatically degrading it.
Consecutive updates of the adaptable weights $\thetabf^{a}$ on narrow portions of the target distribution can cause the model to overspecialize.
Such behavior can be caused by the combination of a sub-optimal choice of hyperparameters for a specific scenario and the lack of sample diversity at the batch level.
Note that the latter does not arise exclusively in video scenarios, but also in situations characterized by a high class imbalance.
Moreover, adapting parameters across the network and within an iterative optimization procedure such as SGD (which spans many batches of data), can inherently lead to the degeneration of the model over time.
To make this more intuitive, in Fig.~\ref{fig:non_iid} we showcase a failure mode of the widely used entropy minimization principle.
In a low intra-batch diversity situation, entropy minimization can degenerate the model \textit{silently}.
In other words, it can fail without exhibiting any distinctive behaviour that, in the absence of labels, would allow for a clear diagnosis.
An illustrative explanation of this phenomenon is conveyed in Fig.~\ref{fig:non_iid_explanation}.
    
\begin{figure*}[t]
	\centering
	\includegraphics[width=0.3\textwidth]{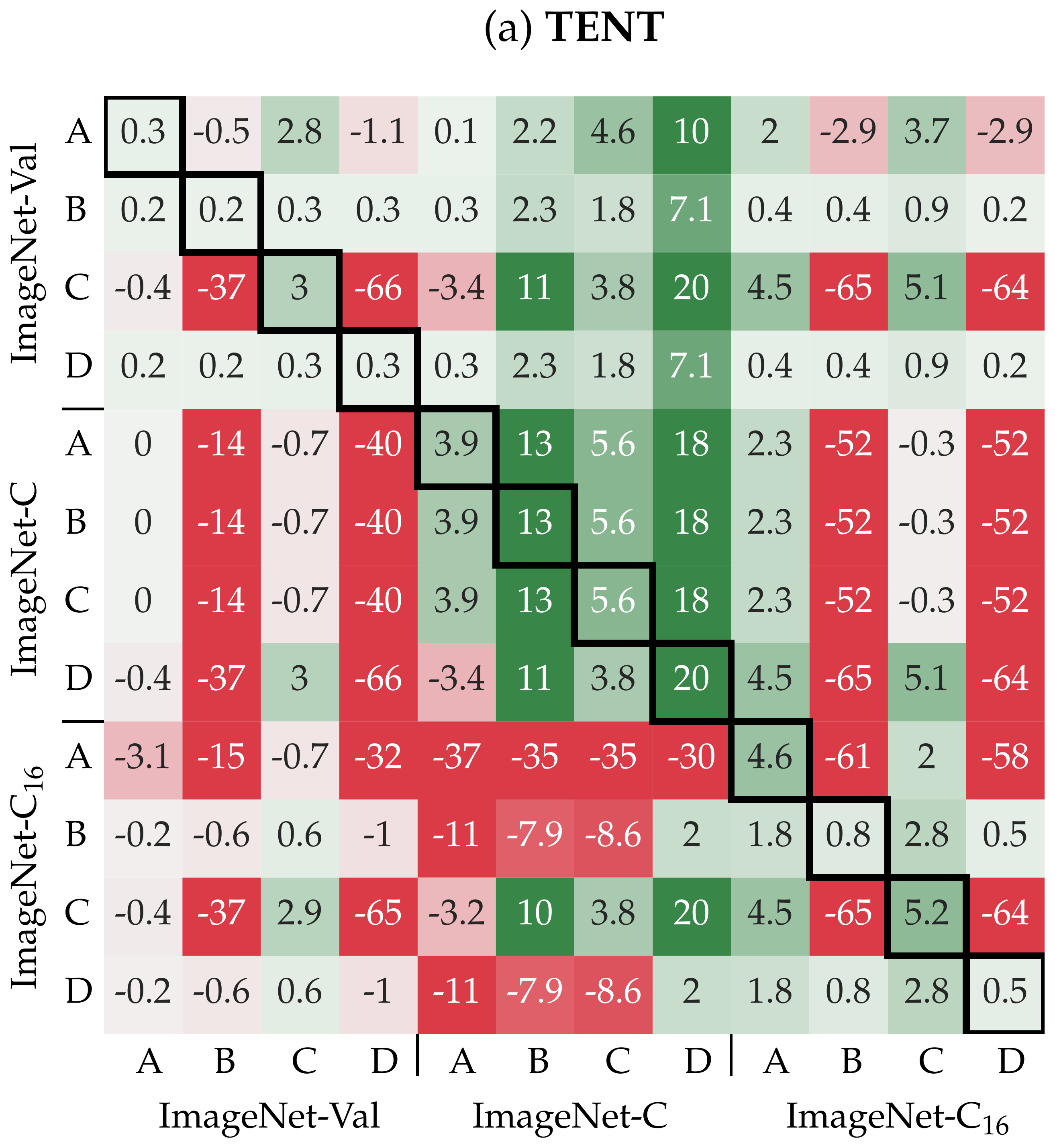} \qquad
	\includegraphics[width=0.32\textwidth]{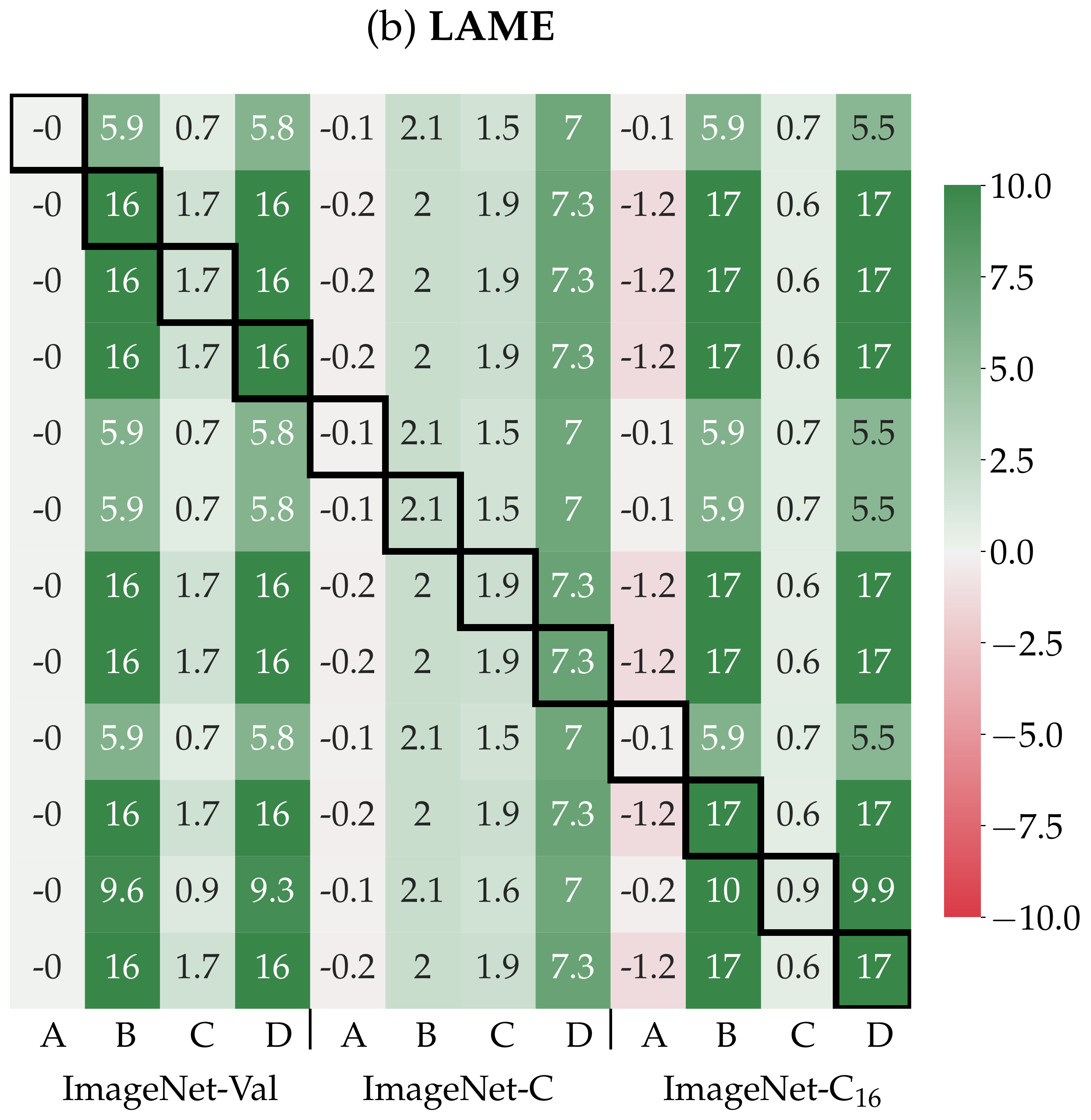}
	\caption{\textit{Cross-shift} validation for TENT \cite{tent} (left) and our proposed LAME (right).
	A cell at position $(i, j)$ shows the absolute improvement (or degradation) of the current method \wrt to the baseline when using the optimal hyperparameters for scenario $i$, but evaluating in scenario $j$.
	Legend: A = i.i.d.,  B = non i.i.d., C = i.i.d. + prior shift, D = non i.i.d. + prior shift. More details on the scenarios in Sec. \ref{sec:experiments}}
	\label{fig:cross_cases}
\end{figure*}

One may argue that choosing optimal hyperparameters may solve the problems mentioned above.
However, tuning hyperparameters separately for each target scenario would require access to the labels.
Moreover, this approach would also require to know which scenario is going be encountered at test time.
These two points defeat the whole purpose of the TTA paradigm.
Therefore, it would be desirable for NAMs' hyperparameters to generalize well \textit{across} scenarios.
However, keeping the entropy minimization approach of TENT~\cite{tent} as an example, we show on the left matrix of Fig.~\ref{fig:cross_cases} that such generalization is, in practice, far from fulfilled.
More specifically, to obtain this matrix, we created a series of 12 validation scenarios (see Section~\ref{sec:experiments}), providing a wide coverage of the shifts discussed in Section \ref{sec:problem_formulation}.
Row $i$ is to be read in the following way: we tune hyperparameters considering only scenario $i$, and then observe to which extent this choice of hyperparameters generalizes to all scenarios $j \in \{1, \dots, 12\}$. 
The absolute improvement (or degradation) \wrt the performance of the non-adapted model is reported in the matrix.
The clear trend emerging from Fig.~\ref{fig:cross_cases} is that the entropy minimization approach is severely brittle \wrt its hyperparameter configuration, especially in non-\iid and class-imbalanced scenarios, where a sub-optimal choice can degrade the model's accuracy by up to an absolute $66\%$ compared to the non-adaptive baseline.
We emphasize that Fig.~\ref{fig:cross_cases} only shows validation results obtained when using scenario-specific hyperparameters, and therefore only serves the purpose of empirically demonstrating the issue with over-specific hyperparameters.
In Appendix, we show that the same trend can be observed for all NAMs we experimented with.

As an alternative, in Section~\ref{sec:method} we propose an adaptation strategy which only affects the output of the model (not its parameters), only considers one batch of data at a time, and has only one hyperparameter to tune.

\section{The \LAME method}
\label{sec:method}
In order to address the aforementioned issues, we introduce a method that only aims at providing a \textit{correction} of the output probabilities of a classifier instead of modifying the internal parameters of its feature extractor.
On the one hand, freezing the source classifier prevents our method from accumulating knowledge across batches.
On the other, it mitigates the risk of degenerating the classifier, reduces compute requirements (as gradients are neither computed nor stored), and inherently removes the need for searching over delicate hyperparameters such as learning rate or momentum of the optimizer. Overall, we empirically demonstrate that such an approach is more reliable and practical than NAMs when the test-time conditions are unknown.

\magicpar{Formulation.}
Assume we are given a batch of data sampled from the target distribution $\XBF \in \mathbb R^{N \times d} \sim p_t^N(\xbf)$, with $N$ the number of samples and $d$ the feature dimension. Our method finds a latent assignment vector $\zbft_i = (\tilde{z}_{ik})_{1 \leq k \leq K} \in \Delta^{K-1}$ for each data point $\xbf_i$, which aims to approximate the true distribution $p(z|\xbf)$, with $K$ the number of classes and $\Delta^{K-1} = \{\zbft \in [0, 1]^K \; | \; {\mathbf 1}^T \zbft  = 1 \}$ the probability simplex. 
A principled way to achieve this is to find assignments $\ZBFT$ that maximize the log-likelihood of the data subject to simplex constraints $\zbft_i \in  \Delta^{K-1}, \ \forall i$:
	\begin{equation} \label{eq:likelihood}
		\mathcal{L}(\ZBFT) = \log \left (\prod_{i=1}^N \prod_{k=1}^K p(\xbf_i, k)^{\tilde{z}_{ik}} \right ) \ceq \sum_{i=1}^N 
		\zbft_i^T \log ({\pbf}_i)
	\end{equation}
where $\ZBFT \in [0, 1]^{NK}$ is the vector that concatenates all assignment vectors $\zbft_i$, ${\pbf}_i = \left (p(k|\xbf_i) \right )_{1 \leq k \leq K} \in \Delta^{K-1}$, and $\ceq$ stands for equality up to an additive constant. In order to prevent over-confident assignments, we consider a negative-entropy regularization that discourages one-hot assignements for $\ZBFT$. Note that such regularization also acts as a barrier that restricts the domain of $\zbft_i$ to non-negative values, hence implicitly handling the $\zbft_i \geq 0$ constraint. Maximizing the regularized log-likelihood objective therefore amounts to minimizing the following Kullback–Leibler (KL) divergences subject to ${\mathbf 1}^T \zbft_i  = 1 , \ \forall i$:


\begin{equation} \label{eq:KL}
	-\sum_{i=1}^N \zbft_i^T \log ({\pbf}_i) + \sum_{i=1}^N \zbft_i^T \log (\zbft_i) = \sum_{i=1}^N \text{KL}(\zbft_i||{\pbf}_i)
\end{equation}
Problem \eqref{eq:KL} is minimized for $\zbft_i = {\pbf}_i, \ \forall i$. Taking a step back, we don't have access to ${\pbf}_i$, but only to the source parametric model
${\qbf}_{i} = (q_{\thetabf}(k|\xbf_i))_{1 \leq k \leq K}$ 
which, recall, might be a poor approximation of the true distribution when evaluated on target samples $\xbf \sim p_t(\xbf)$. In fact, simply replacing ${\pbf}_i$ by 
${\qbf}_{i}$ 
in Eq.~\eqref{eq:KL} yields the predictions from the source model as optimum: 
$\zbft_i = \qbf_{i}$. 

To compensate for the inherent error of this approximation,
we focus on Laplacian regularization, which encourages neighbouring points in the feature space to have consistent latent assignments.
Laplacian regularization is widely used in semi-supervised learning \cite{belkin2006manifold,Chapelle2010,IscenTAC19}, where it is optimized jointly with supervised losses over labelled data points, or in graph clustering \cite{shiMalik2000,shaham2018spectralnet,Tang2019KernelCK}, where it is optimized subject to class-balance constraints.
The TTA problem is different as, unlike semi-supervised learning, cannot count on any supervision and, unlike clustering, class-balance constraints are irrelevant (or even detrimental).
Hence, we introduce Laplacian Adjusted Maximum-likelihood Estimation (\LAME), which 
minimizes the likelihood in \eqref{eq:KL} jointly with a Laplacian correction, subject to constraints ${\mathbf 1}^T \zbft_i  = 1, \ \forall i$:
\begin{equation}
\label{LAME}
\mathcal{L}^{\text{LAME}}(\ZBFT) = \sum_i \text{KL}(\zbft_i||{\qbf}_i) - \sum_{i,j} w_{ij} \zbft_i^T \zbft_j
\end{equation}
where $w_{ij} = w(\phi(\xbf_i), \phi(\xbf_j))$, with $\phi$ denoting our pre-trained feature extractor and $w$ is a function 
measuring the affinity between $\phi(\xbf_i)$ and $\phi(\xbf_j)$. The closer the points in the feature space, the higher their affinity. Clearly, when the affinity is high ($w_{ij}$ is large), minimizing the Laplacian term in \eqref{LAME} seeks the largest possible value of dot product $\zbft_i^T \zbft_j$, thereby assigning points $i$ and $j$ to the same class. Therefore, our model in \eqref{LAME} could be viewed as a graph clustering of the batch data, penalized by a KL term discouraging substantial deviations from the source-model predictions.

\begin{figure*}[t]
	\includegraphics[width=\textwidth]{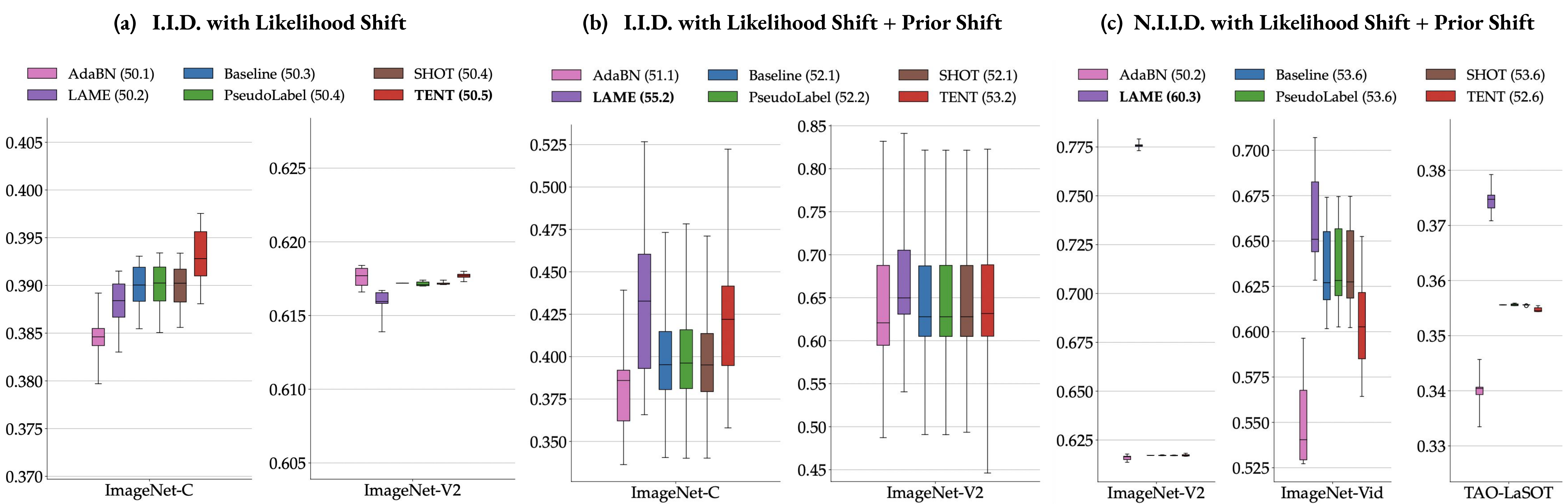}
\caption{Results across the 7 testing scenarios, using the same Original RN-50 that was used for validation. The average for each scenario is reported in the legend. The batch size is 64. Each experiment is run 10 times with different random seeds. Experiments with \textit{prior shift} tend to exhibit larger variance, since each random run uses new class proportions, each sampled from a Zipf distribution.}
	\label{fig:test_boxes}
\end{figure*}

\magicpar{Efficient optimization via a concave-convex procedure.}
In what follows, we show that our Problem~\eqref{LAME} can be minimized using the Concave-Convex Procedure (CCCP) \cite{Yuille2001}, which allows us to obtain a highly efficient iterative algorithm, with convergence guarantee. 
Each iteration updates the current solution $\ZBFT^{(n)}$ as the minimum of a tight upper bound on the objective.
This guarantees that the objective does not increase at each iteration.
For the sum of a concave and a convex function, as is the case of our objective in \eqref{LAME}, a CCCP replaces the concave part by its linear first-order approximation at the current solution, which is a tight upper bound, while keeping the convex part unchanged.
In our case, the Laplacian term is concave when the affinity matrix $W = [w_{i,j}]$ is positive semi-definite, 
while the KL term is convex.
The concavity of the Laplacian for positive semi-definite $W$ could be verified by re-writing the term as follows\footnote{$W$ positive semi-definite implies $W \otimes I$ positive semi-definite.}: $- \sum_{i,j} w_{ij} \zbft_i^T \zbft_j = - \ZBFT^T (W \otimes I) \ZBFT$, where $\otimes$ denotes the Kronecker product and $I$ is the $K$-by-$K$ identity matrix. 
We thus replace the Laplacian term in \eqref{LAME} by $-((W \otimes I) \ZBFT^{(n)})^T \ZBFT$, which yields the following tight upper bound, up to an additive constant independent of $\ZBFT$:
\begin{equation}
	\label{LAME-bound}
	\mathcal{L}^{\text{LAME}}(\ZBFT) \cleq \sum_i \text{KL}(\zbft_i||{\qbf}_i) - ((W \otimes I) \ZBFT^{(n)})^T \ZBFT  
\end{equation}
Solving the Karush-Kuhn-Tucker (KKT) conditions corresponding to minimizing convex upper bound
\eqref{LAME-bound}, subject to constraints ${\mathbf 1}^T \zbft_i  = 1, \ \forall i$, yields the 
following decoupled updates of the assignment variables:
\begin{align}
	\label{final-updates}
	\tilde{z}_{ik}^{(n+1)} = \frac{q_{\thetabf}(k|\xbf_i) \exp(\sum_j w_{ij} \tilde{z}_{jk}^{(n)})}{\sum_{k'} q_{\thetabf}(k'|\xbf_i) \exp(\sum_j w_{ij} \tilde{z}_{jk'}^{(n)})}
\end{align}
which have to be iterated until convergence. The full derivation of Eq. \eqref{final-updates} is provided in Appendix.

\section{Experimental design}
\label{sec:experiments}

The design of our experimental protocol is mainly guided by the desire to assess both model and domain independence of TTA methods.
For model independence, we need to evaluate the performance of methods under a variety of pre-trained models.
As for domain independence, a single fixed trained model must allow to evaluate a TTA method under multiple adaptation scenarios.
This implies that the source classes encoded in the pre-trained model must be able to adequately cover the classes of interest that may be encountered at test time.
Note that, in practice, this is a reasonable requirement, as modern large-scale datasets span tens of thousands of classes~\cite{ridnik2021imagenet,imagenet,OpenImages,wu2019tencent}.

\magicpar{Networks.} 
Because of their popularity within the community and the large number of classes covered, ImageNet-trained models represent an ideal playground for our experiments.
In particular, they allow to evaluate model independence along two axis.
First, with respect to the training procedure, by experimenting with the same ResNet-50 architecture (RN-50 herein), but trained in three different ways: the original release from Microsoft Research Asia (MSRA)~\cite{resnet}, Torchvision's~\cite{pytorch}, and using the self-supervised SimCLR~\cite{chen2020simple}.
Second, with respect to the architecture itself, by providing results on 5 different backbones, including RN-18, RN-50, RN-101, EfficientNet (EN-B4)~\cite{efficientnet} and the recent Vision Transformer ViT-B~\cite{vit_dosovitskiy2020image}. All models used were trained on the standard ImageNet ILSVRC-12 training set, except for ViT-B which uses an additional ImageNet-21k \cite{imagenet_21k} pre-training step.

\begin{figure}
	\centering
	\includegraphics[width=0.45\textwidth]{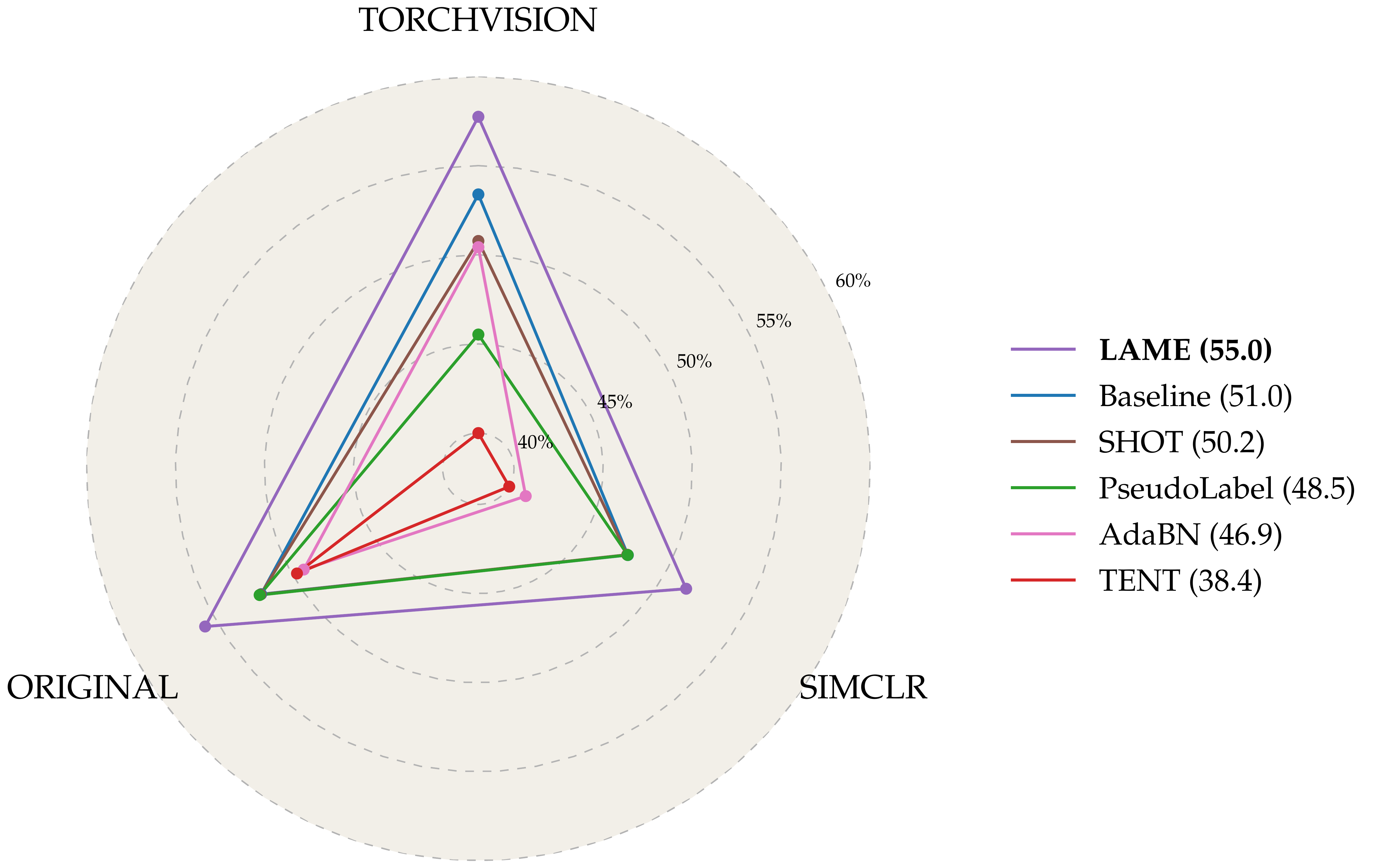}   \\
	\hspace{5em}
	\includegraphics[width=0.45\textwidth]{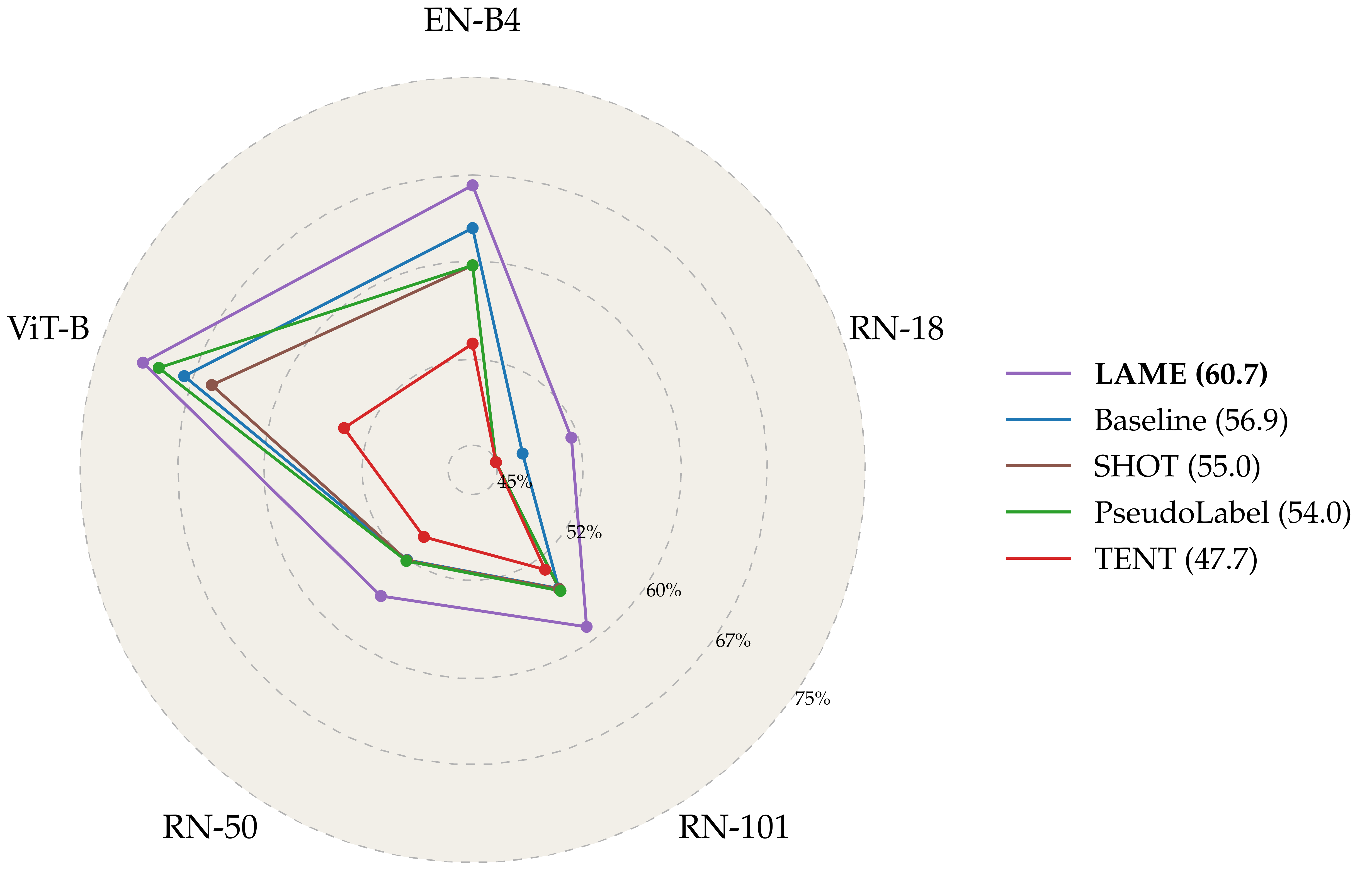}
	\caption{\textbf{Transferability of hyperparameters across models}.
	For each TTA method, we use the optimal set of hyperparameters obtained during validation and using the original release of RN-50~\cite{resnet} as backbone. Each vertex on the chart represents the average across our 7 test scenarios for a specific architecture (exact values available in Table \ref{tab:result}). The values in the legend represent the average over all the vertices.
	(Top): We test these hyperparameters using the same backbone but different training procedures. \textit{Torchvision} refers to the RN-50 checkpoint available in PyTorch's model zoo, SimCLR to the checkpoint obtained from the self-supervised approach of~\cite{chen2020simple}, and \textit{original} refers to the same model used to choose the hyperparameters.
	(Bottom): The \textit{same} set of hyperparameters is used for different architectures, ranging from a RN-18 to the recent vision transformer ViT-B~\cite{vit_dosovitskiy2020image}.
	To allow similar setups across architectures, a batch size of 16 is used to generate the above results.}
	\label{fig:hparams_robustness}
\end{figure}

\magicpar{Hyperparameter search.}  

\begin{table}
    \centering
    \resizebox{0.4\textwidth}{!}{
        \begin{tabular}{lcccc}
            \toprule
            Method & \textit{Original} & \textit{Torchvision} & \textit{SimCLR} & Mean\\
            \midrule
            Baseline & 52.07 & 53.39 & 47.68 & 51.0 \\
            PseudoLabel  \cite{lee2013pseudo} & 52.17 & 45.53 & 47.67 & 48.5 \\
            SHOT-IM \cite{shot} & 52.11 & 50.77 & 47.64 & 50.2\\
            TENT \cite{tent} & 49.74 & 28.54 & 37.00 & 38.4\\
            AdaBN \cite{ada_bn} & 49.31 & 50.43 & 41.07 & 46.9\\
            \rowcolor{Gray} LAME &  \textbf{55.70} & \textbf{57.74} & \textbf{51.46} & \textbf{55.0}\\
            \bottomrule
        \end{tabular}
    } \\
    \vspace{1em}
    \resizebox{0.49\textwidth}{!}{
        \begin{tabular}{lcccccc}
            \toprule
            Method & RN-18 & RN-50 & RN-101 & EN-B4 & ViT-B/16 & Mean\\
            \midrule
            Baseline & 47.28 & 52.07 & 54.93 & 62.67 & 67.69 & 56.9 \\
            PseudoLabel \cite{lee2013pseudo} & 33.13 & 52.17 & 55.17 & 59.64 & 69.86 & 54.0 \\
            SHOT-IM \cite{shot} & 42.95 & 52.11 & 54.98 & 59.64 & 65.33 & 55.0 \\
            TENT \cite{tent} & 28.59 & 49.74 & 53.04 & 53.27 & 54.00 & 47.7 \\
            \rowcolor{Gray} LAME & \textbf{51.46} & \textbf{55.70} & \textbf{58.79} & \textbf{66.17} & \textbf{71.22} & \textbf{60.7} \\
            \bottomrule
        \end{tabular}
    }
    \caption{Numeric values of Fig. \ref{fig:hparams_robustness} for easier referencing. Each number represents a vertex in Fig. \ref{fig:hparams_robustness} denoting the average performance across our 7 test scenarios using some backbone.}
    \label{tab:result}
\end{table}

For validation purposes, we consider 3 datasets. First, we use the original validation set of ImageNet \cite{imagenet}. To represent \textit{likelihood shift}, we consider \textit{ImageNet-C-Val}, which augments the original images with 9 realistic perturbations of varying intensity (the other 10 from the original ImageNet-C \cite{imagenet_c} are reserved for testing). Finally, we consider \textit{ImageNet-C$_{16}$}, a variant of \textit{ImageNet-C} that simulates an easier but practical scenario where a subset of ImageNet classes is mapped to 16 superclasses. By reducing the total number of classes, ImageNet-C$_{16}$ also reduces class diversity at the batch level, which we identified as a potentially critical factor for NAMs approaches in Section \ref{sec:network_adaptation}. In order to mimic realistic prior shifts, we modify the class ratios to follow a Zipf distribution~\cite{reed2001pareto}. Finally, to cover non-\iid scenarios, we present the model with a sequence of ``tasks'', where each task either represents a set of samples perturbed with the same corruption (in the case of ImageNet-C), or belonging to the same class otherwise. All the combinations of 3 datasets, 2 prior shifts (with and without Zipf-unbalanced class distribution) and 2 sampling schemes (\iid or non-\iid) add up to the 12 validation scenarios.
For each method, a grid-search over salient hyperparameters is carried out, and the single hyperparameter set that obtains best average performance over the 12 validation scenarios is selected, and \textbf{kept fixed for test experiments in Fig.~\ref{fig:test_boxes} and \ref{fig:hparams_robustness}}. The exact definition of the grid-search for each method is available in the Appendix.

\magicpar{Testing.}  For testing, we design 4 \iid and 3 non-\iid test scenarios.
For the \iid cases, we use the 4 combinations obtained by coupling ImageNet-C-Test and ImageNet-V2 \cite{imagenet_v2} with the presence or absence of Zipf class-imbalance.
As for the 3 non-\iid. scenarios, we use again ImageNet-V2 (with a different split), along with two video datasets: ImageNet-VID \cite{imagenet} and the LaSOT subset from TAO~\cite{tao}.
Keeping the idea of feeding the model with a sequence of tasks, video datasets allow us to evaluate realistic scenarios by simply grouping frames from the same video together.
We use 10 random runs for each test experiment.
More details on all datasets (and class mappings) in Appendix.

\magicpar{Methods.}
As a first baseline, we evaluate the source-trained model without any adaptation, referred to as \textit{Baseline}.
For Network Adaptation Methods (NAMs), we reproduce and evaluate four state-of-the-art TTA methods that can be run in an online fashion: \textit{TENT} \cite{tent} based on entropy minimization, \textit{SHOT-IM} based on mutual information maximization, \textit{PseudoLabel} \cite{pseudo_label} based on min-entropy minimization and \textit{AdaBN} \cite{ada_bn} based on batch normalization statistics alignment.
Finally, we evaluate \LAME.

\section{Experimental results} \label{expm:results}

\magicpar{Towards domain-independent test-time adaptation.}
As motivated in Section \ref{sec:network_adaptation}, most scenario-sensitive hyperparfameters come from the optimization of the network.
By virtue of completely freezing the classifier, our \LAME approach is free of such burden.
Instead, \LAME only tries to find optimal shallow assignments through a bound-optimization procedure that does not introduce any hyperparameter.
Therefore, we are only left with the tuning of the affinity function $w$ from Eq. \eqref{LAME}, which is less sensitive than the optimization-related hyperparameters of NAMs.
This claim is first supported by inspecting \LAME's \textit{cross-shift} validation matrix, already used earlier to illustrate NAMs' brittleness. Looking this time at the right plot of Fig.~\ref{fig:cross_cases}, we can see drastic improvements both in terms of average performance and worst-case degradation across all cases \wrt TENT.
\footnote{We speculate that introducing more hyperparameters in \LAME (\eg weighting the different terms of our loss) would result in worse off-diagonal terms in Fig.~\ref{fig:cross_cases}, but also higher overall performance.}
A second empirical evidence supporting this claim comes from the results on the test scenarios, shown in Fig.~\ref{fig:test_boxes}. Consistent with the validation results in Fig.~\ref{fig:cross_cases}, Fig.~\ref{fig:test_boxes} confirms that \LAME does not help in standard \iid \emph{likelihood shifts}, and fares around $0.5 \%$ below the baseline in worst cases.
However, when \emph{prior shifts} are introduced, NAMs' performance does not improve over the baseline, whereas \LAME exhibits very noticeable improvements.
This is particularly evident in non-\iid scenarios, where the average improvement is of (absolute) $6.7\%$, and goes up to $15\%$ in the case of ImageNet-v2.
Note that such improvement comes almost independently of the batch size used, as shown in Appendix. 


\magicpar{NAMs are brittle \wrt the training procedure.}
As for model-independence, we first inspect whether methods are robust to changes to the training procedure.
Such robustness is desired, for instance, in the case where the provider of the source model has released an update:
in such a case, a TTA method should not require a new round of validation.
As a first scenario, we investigate whether the set of hyperparameters obtained using the Original RN-50~\cite{resnet} generalizes to the same methods, but when using the RN-50 provided by Torchvision.
Given that both models were trained with standard supervision and minor experimental differences, one would expect
the optimal set of hyperparameters to be very similar in the two cases.
Results on the top chart of Fig.~\ref{fig:hparams_robustness} suggest quite the opposite.
While \LAME preserves the same improvement \wrt the baseline, all NAMs lose significant ground,
with TENT performing particularly poorly.
We further experiment with a RN-50 trained using the self-supervised \textit{SimCLR}, and observe that \LAME once again retains its relative improvement of 4\% \wrt the baseline, with no other method beating it.

\magicpar{LAME generalizes across architectures, NAMs don't.}
Generalizing across different architectures should be a desirable property for any TTA method.
In particular, for very large models, an exhaustive validation can become prohibitively expensive, thus making ``model plug-and-play'' an attractive feature.
Results using five architectures (EfficientNet-B4~\cite{efficientnet}, the three ResNet variants, and the larger ViT-B~\cite{vit_dosovitskiy2020image} transformer) are shown on the bottom chart of Fig.~\ref{fig:hparams_robustness}.
Across the board, LAME is the only method able to retain a consistently significant improvement \wrt the baseline, which remains a better option than any of the NAMs, especially with small backbones such as RN-18.

\begin{figure}[h!]
	\centering
	\includegraphics[width=0.65\columnwidth]{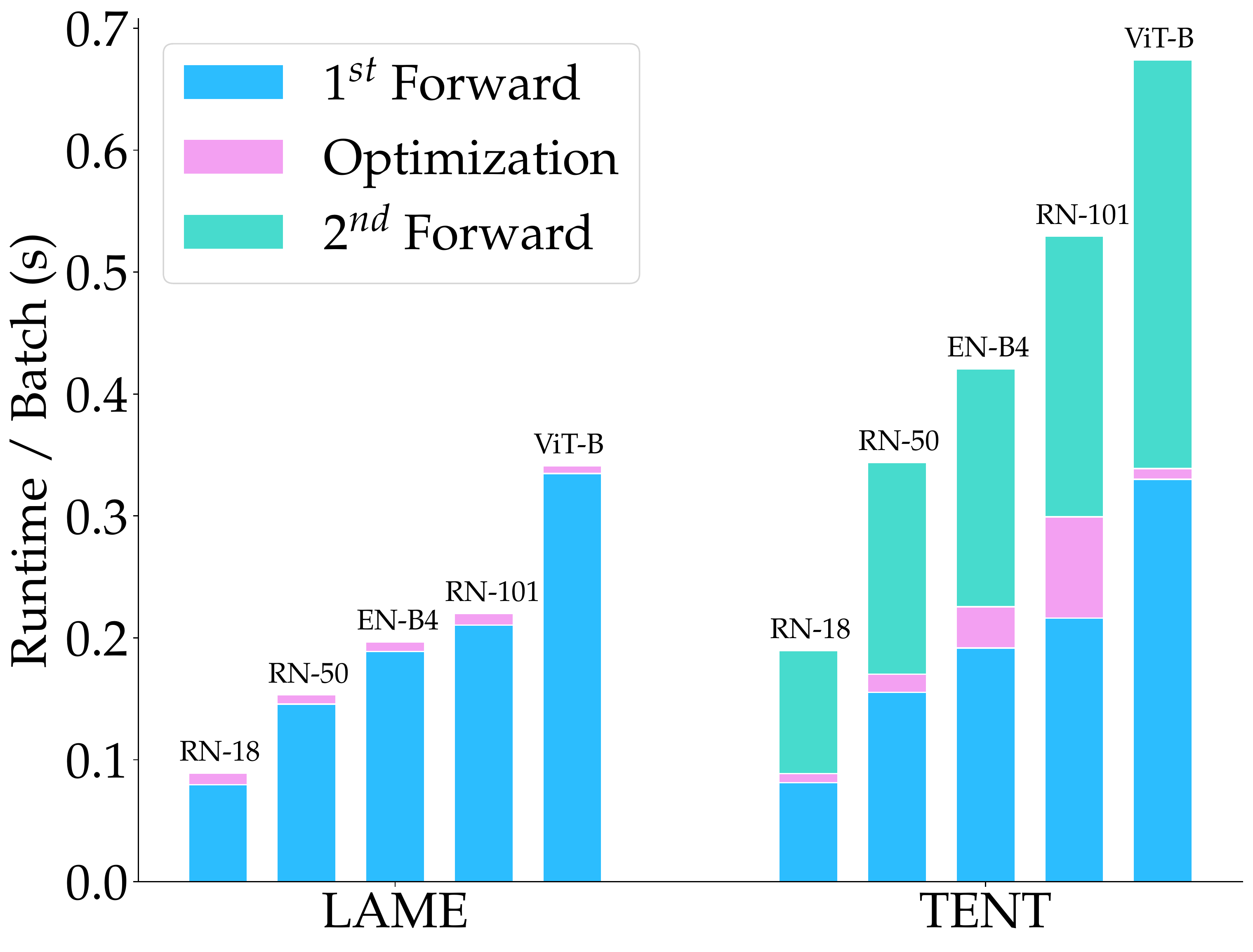}
	\caption{\textbf{Runtime per batch} of LAME vs TENT for 5 different backbones: RN-18, RN-50, EN-B4, RN-101 and ViT-B.
	Batch 64 is used for the RN-* family, and 16 for EN-B4 and ViT-B (as both use 380x380 images instead of 224x224). \LAME provides corrected outputs without requiring a second forward pass.}
	\label{fig:runtimes}
\end{figure}

\magicpar{LAME runs twice as fast, while requiring twice less memory than NAMs.}
Provided that several direct applications of test-time adaptation involve real-time adaptation to data streams, the ability to run as efficiently as possible can also be a critical factor for practitioners.
To measure runtimes, we divide inference into 3 stages: $1^{st}$ forward, optimization (corresponding to SGD for NAMs and to the bound-optimization procedure of Section~\ref{sec:method} for \LAME), and $2^{nd}$ forward (only needed for methods that modify the parameters of the model). 
Altogether, these three contributions account for the total runtime of each method.
Results provided in Fig.~\ref{fig:runtimes} testify the clear advantage of \LAME over the representative TENT (runtimes of other NAMs were found roughly similar to TENT).
Memory-wise, \LAME does not require to keep any gradient or intermediary buffer, which roughly halves the amount of GPU memory needed \wrt NAMs.

\label{expm:results}

\section{Conclusion}
\label{sec:conclusion}

Motivated by the high cost of training new models, we proposed a novel approach for online test-time adaptation (TTA) that is agnostic to both training and testing conditions.
We introduced an extensive experimental protocol covering several datasets, realistic shifts and models, and evaluated existing TTA approaches by making sure that test-time domain information would not leak to inform the hyperparameters' choice.
Across the board, these methods underperform a non-adaptive baseline and can even lead to a catastrophic degradation of performance.
We identified over-adaptation of the model parameters
as a strong suspect for the poor performance of these methods, and opted for a more conservative approach that only corrects the output of the model. 
We proposed Laplacian Adjusted Maximum-likelihood Estimation (LAME), an unsupervised objective that finds the optimal set of latent assignments by discouraging deviations from the prediction of the pre-trained model, while at the same time encouraging label propagation under the manifold smoothness assumption.
Averaging accuracy over the many scenarios considered, LAME outperfoms all existing methods and the non-adaptive baseline, while requiring less compute and memory.
Nonetheless, being restricted to the classifier's output, LAME is also inherently limited.
For one, it does not noticeably help in standard \iid and class-balanced scenarios.
We hope that our work will motivate further developments in this line of research.
In particular, we believe that methods adopting a hybrid adaptation/correction approach, if choosing their hyperparameters under a strict regime, will have the potential to effectively tackle an even wider variety of scenarios.

{
\small
\bibliographystyle{ieee_fullname}
\bibliography{main_cvpr22}
}

\clearpage
\appendix

\section{Mapping between source and target classes} \label{sec:formal_link_z_y}

As stated towards the end of Section \ref{sec:problem_formulation}, target classes may belong to a set of \textit{superclasses}. Formally, we note original source classes as $\mathcal{Y}$, and target classes as $\mathcal{Z}$. We require that there exists a mapping $\mu: \mathcal{Y} \rightarrow \mathcal{Z} \cup \{\emptyset \}$ that maps each source class to either its unique corresponding superclass in the target domain if it exists, or to the null variable. 


\begin{figure}[h]
	\centering
	\includegraphics[scale=0.4]{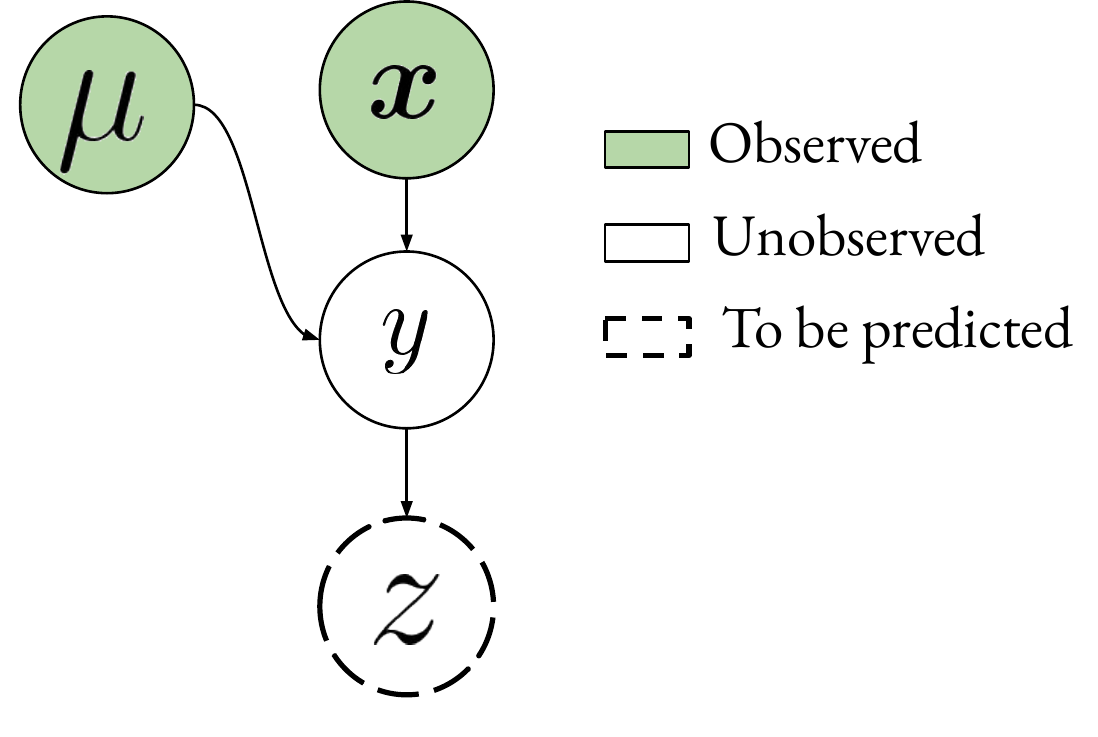}
	\caption{A causal view of the class mapping. }
	\label{fig:causal_graph}
\end{figure}

In preliminary experiments, we explored max-pooling following \cite{taori2020measuring}:
\begin{align}
	q_{\thetabf}(z|\mathbf{x}, \mu) = \max_{y: \mu(y)=z} q_{\thetabf}(y|\mathbf{x})
\end{align}
but found that average pooling performed overall better:
\begin{align}
	q_{\thetabf}(z|\mathbf{x}, \mu) = \frac{1}{|\{y: \mu(y)=z\}|} \sum_{y: \mu(y)=z} q_{\thetabf}(y|\mathbf{x})
\end{align}







\section{Detailed derivation of \LAME}

	We detail the derivation of Eq. \eqref{final-updates} from the upper bound \eqref{LAME-bound}. Given a solution $\ZBFT^{(n)}$ at iteration $n$, the goal is find the next iterate $\ZBFT^{(n+1)}$ that minimizes the following constrained problem:

	\begin{align} \label{eq:ub_problem}
		\min_{\ZBFT} & \quad \sum_{i=1}^N \text{KL}(\zbft_i||\qbf_i) - \sum_{i=1}^N \sum_{j=1}^N \zbft_i^T \zbft_j^{(n)} \\
		\text{s.t}   & \quad \zbft_i^T \ones_K, \ \forall i \in \{1, \dots, N\} \nonumber
	\end{align}

	The objective function of \eqref{eq:ub_problem} is strictly convex due to the presence of the KL term, with linear equality constraints. We can write down the associated Lagrangian:

	\begin{align}
		\mathcal{L}(\ZBFT, \lbf) =& \sum_{i=1}^N \zbft_i^T \log(\zbft_i) - \sum_{i=1}^N \zbft_i^T \log(\qbf_i)  \\ 
								  & - \sum_{i=1}^N \sum_{j=1}^N \zbft_i^T \zbft_j^{(n)} + \sum_{i=1}^N \lbf_i(\zbft_i^T \ones_K - 1) \nonumber
	\end{align}
	where $\lbf=\{\lambda_1, \dots, \lambda_N\}$ represents the vector of Lagrange multipliers associated with the N linear constraints of Problem \eqref{eq:ub_problem}. Let us now compute the derivative of $\mathcal{L}(\ZBFT, \lbf)$ w.r.t to $\zbft_i, \ \forall i \in \{1, \dots, N \}$:

	\begin{align} \label{eq:lagrange_gradient}
		\nabla_{\zbft_i} \mathcal{L}(\ZBFT, \lbf) = (1 + \lambda_i)\ones_K + \log(\zbft_i) - \log(\qbf_i) - \sum_{j=1}^N w_{ij} \zbft_j^{(n)}
	\end{align}

	By setting the gradients of \eqref{eq:lagrange_gradient} to 0, we can obtain for the optimal solution $\zbft_i^{(n+1)}$:

	\begin{align} \label{eq:temp_solution}
		\zbft_i^{(n+1)} = (\qbf_i \odot \exp(\sum_{j=1}^N w_{ij} \zbft_j^{(n)})) \exp(-(\lambda_i + 1))
	\end{align}

	Combining Eq. \eqref{eq:temp_solution} with the constraint $\ones_K^T \zbft_i^{(n+1)} = 1$ allows us to recover the Lagrange multiplier:

	\begin{align}
		\lambda_i = \log((\qbf_i \odot \exp(\sum_{j=1}^N w_{ij} \zbft_j^{(n)}))^T \ones_K) - 1
	\end{align}

	Which leads to the final solution:
	\begin{align}
		\zbft_i^{(n+1)} = \frac{\qbf_i \odot \exp(\sum_{j=1}^N w_{ij} \zbft_j^{(n)}))}{(\qbf_i \odot \exp(\sum_{j=1}^N w_{ij} \zbft_j^{(n)}))^T \ones_K}
	\end{align}

\section{Hyperparameters}

	Given that the space of hyperparameters grows exponentially in the dimension of the grid, we are forced to make a decision about which subset of hyperparameters to tune, and which one to keep fixed \wrt the original methods.
	Note that, for all NAMs, we adopt the Adam optimizer, as it can also be easily used without any momentum (which was found to be the best choice for non-\iid cases).	
	




	We define a common grid-search for all NAMs along the following axes:

	\magicpar{Learning rate.} 
		The learning rate plays a crucial in the learning dynamic. In particular, too small a learning rate can prevent NAMs for actually improving the model, while too aggressive ones may completely degenerate the model, as shown in Fig. \ref{fig:non_iid}. Therefore, we search over three values $\{0.001, 0.01, 0.1\}$.

	\magicpar{Optimization momentum.} Although not often optimized for, we found the presence of momentum could heavily degrade the performances, especially in non \iid scenarios, as sharp changes in the distribution violate the underlying data distribution smoothness taken by the presence of momentum. We therefore leave the choice between the standard momentum value of $0.9$ and no momentum at all $0$.

	\magicpar{Batch Norm momentum.} TENT \cite{tent} uses the statistics of the current batch for standardization in Batch Normalization (BN) layers, instead of those from the source distribution computed over the whole source domain. AdaBN \cite{ada_bn} method also relies on target samples' statistics to improve the performances, and so does SHOT \cite{shot} (by using the model in the default training mode with a BN momentum of 0.1).
	While helping in some scenarios, we observed that the use of target statistics in BN normalization procedure can sometimes degrade the results, which echoes the recent findings in \cite{burns2021limitations}. Therefore, we leave methods the choice to only use the statistics from the source domain (\texttt{momentum=0}), the statistics from the current batch (\texttt{momentum=1}), or a trade-off that allows to update the statistics in a smooth manner (\texttt{momentum=0.1}).

	\magicpar{Layers to adapt.} Following the interesting findings from authors in \cite{burns2021limitations}, who showed that adapting only the early layers of a network could greatly help in tackling \textit{prior shifts}, we search over three rough partitions of the set of layers: Adapting the first half of the network while keeping the rest frozen, adapting the second half while keeping the first half frozen, or adapting the full network.
	
	\subsection{Hyperparameters for \LAME}
	
\begin{figure}
	    \centering
	    \includegraphics[width=\columnwidth]{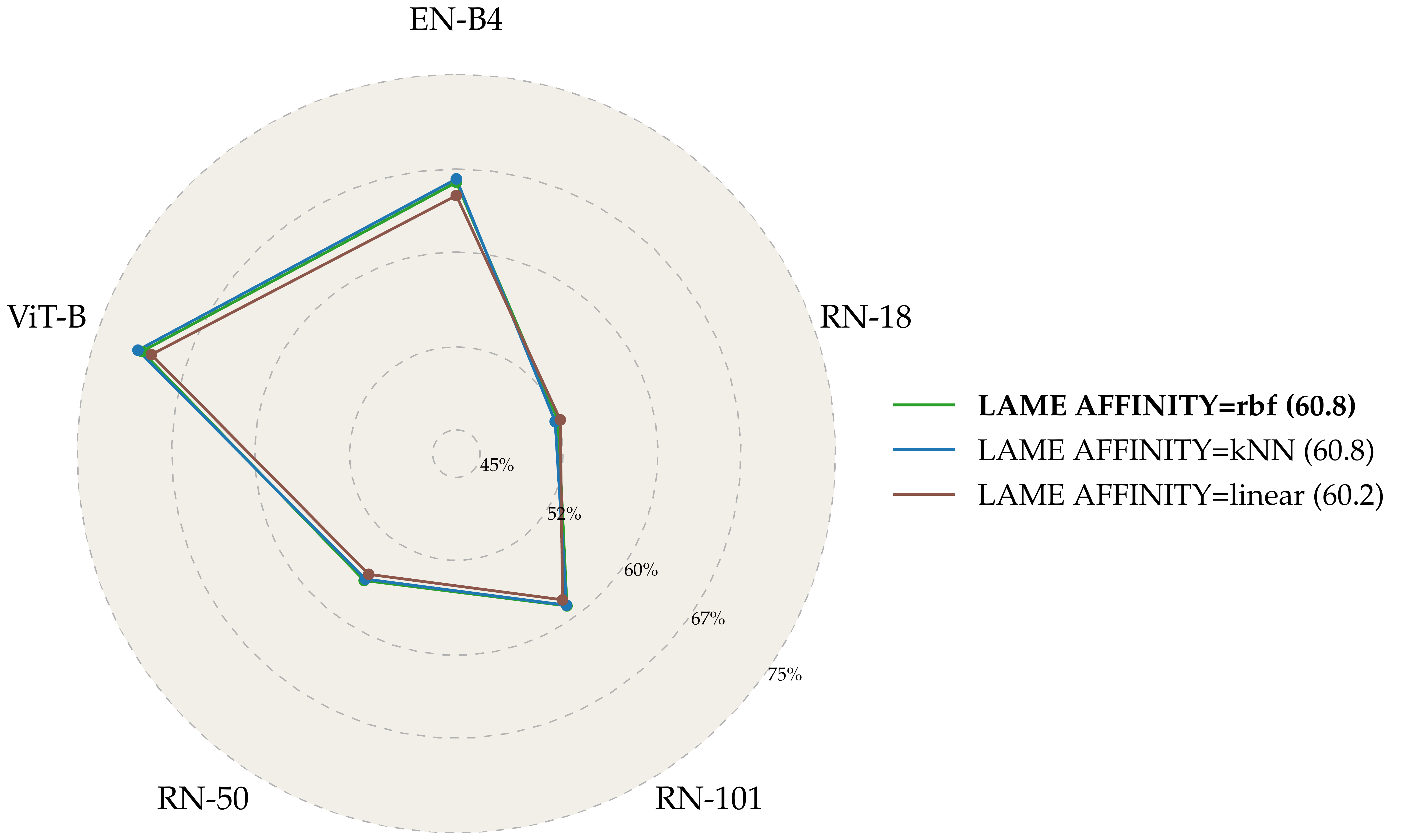}
	    \caption{A comparison of three kernels for \LAME method. The specific choice of the kernel doesn't seem to have much influence on the performances.}
	    \label{fig:kernel_comparison}
	\end{figure}

	Given that $\LAME$ considers the network fully frozen, all hyperparameters detailed above do not need to be tuned for. Instead, \LAME's only hyperparameters are to be found in the choice of the affinity matrix. In this work, we decided to follow a standard choice made in \cite{ziko2020laplacian} to use a $k$-NN affinity, where:
	
	\begin{align}
		w(\phi(\xbf_i), \phi(\xbf_j)) =     \begin{cases}
                        					      1 & \text{if $\phi(\xbf_i) \in \text{kNN}(\phi(\xbf_j))$}\\
                        					      0 & \text{otherwise} \\
                        				    \end{cases}    
	\end{align}
	
	where kNN$(.)$ is the function that returns the set of $k$ nearest neighbours. Therefore, one only needs to tune the value of $k$, which was selected among $\{1, 3, 5\}$. Note that we tried with other standard kernels, namely the simple linear kernel $w(\phi(\xbf_i), \phi(\xbf_j))=\phi(\xbf_i)^T\phi(\xbf_j)$ and the radial kernel $w(\phi(\xbf_i), \phi(\xbf_j))=\exp(-\frac{\norm{\phi(\xbf_i) - \phi(\xbf_j)}^2}{2\sigma^2})$, with $\sigma$ chosen as the average distance of each point to its $k^{th}$ neighbour, and $k$ found through validation. A comparison over the 5 model architectures used is provided in Fig. \ref{fig:kernel_comparison}, where each vertex indicates the average accuracy over the 7 test scenarios. Overall, \LAME equipped with any of the three kernels $\{$kNN, linear, rbf$\}$ performs roughly similarly.
	
	\begin{table*}[t]
	\centering
	\caption{Summary of the different datasets used.}
	\label{tab:datasets}
	\begin{tabularx}{\textwidth}{lcXccc}
		Used for & Dataset & Short description & \# classes & \# samples \\ 
		\toprule
		\multirow{5}{*}{Validation} & ImageNet-Val & The original validation set of ILSVRC 2012 challenge 
		\cite{imagenet}. & 1000 & 50'000 \\
		& ImageNet-C-Val & Corrupted version of ImageNet-Val, with 9 different corruptions used. & 1000 & 450'000 \\
		& ImageNet-C$_{16}$ & Smaller version of ImageNet-C-Val, where only 32 out of the 1000 source classes are mapped to 16 superclasses. & 16 & 14'400 \\
		\midrule
		\multirow{15}{*}{Testing} &  ImageNet-V2 & A recently proposed validation set of ImageNet dataset, collected independently from the original validation set, but following the same protocol. Only 10 samples per class instead of 50 in the original validation set. On ImageNet-V2, models surprisingly drop by $\approx 10 \%$ \wrt their performance on the original ImageNet-Val. & 1000 & 10'000 \\
		& ImageNet-C-Test & Corrupted version of ImageNet-Val, with 10 different corruptions used. & 1000 & 500'000 \\
		& ImageNet-Vid & Video dataset corresponding to the validation set of the full ImageNet-Vid \cite{imagenet}. A collection of videos covering diverse situations for factors such as movement type, level of video clutterness, average number of object instance, and several others. & 30 & 176'126 \\
		& TAO-LaSOT & LaSOT subset of TAO dataset \cite{tao}, a popular tracking benchmark, where each sequence comprises various challenges encountered in the wild. & 76 & 6'558  \\
		\bottomrule
	\end{tabularx}
\end{table*}
    
\section{Cross-shift validation matrices}

\begin{figure*}[t]
	\centering
	\includegraphics[width=0.35\textwidth]{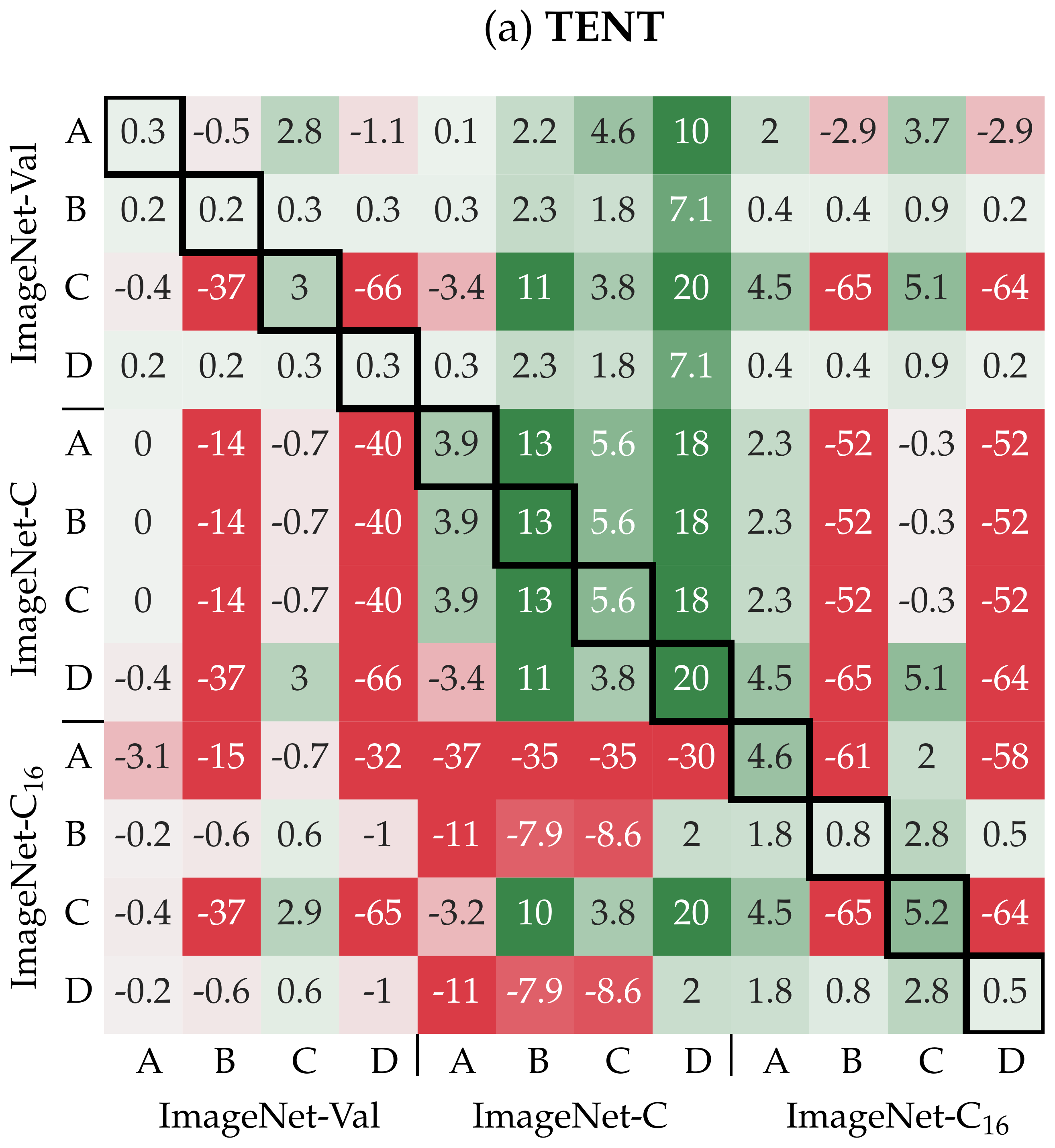} \qquad
	\includegraphics[width=0.35\textwidth]{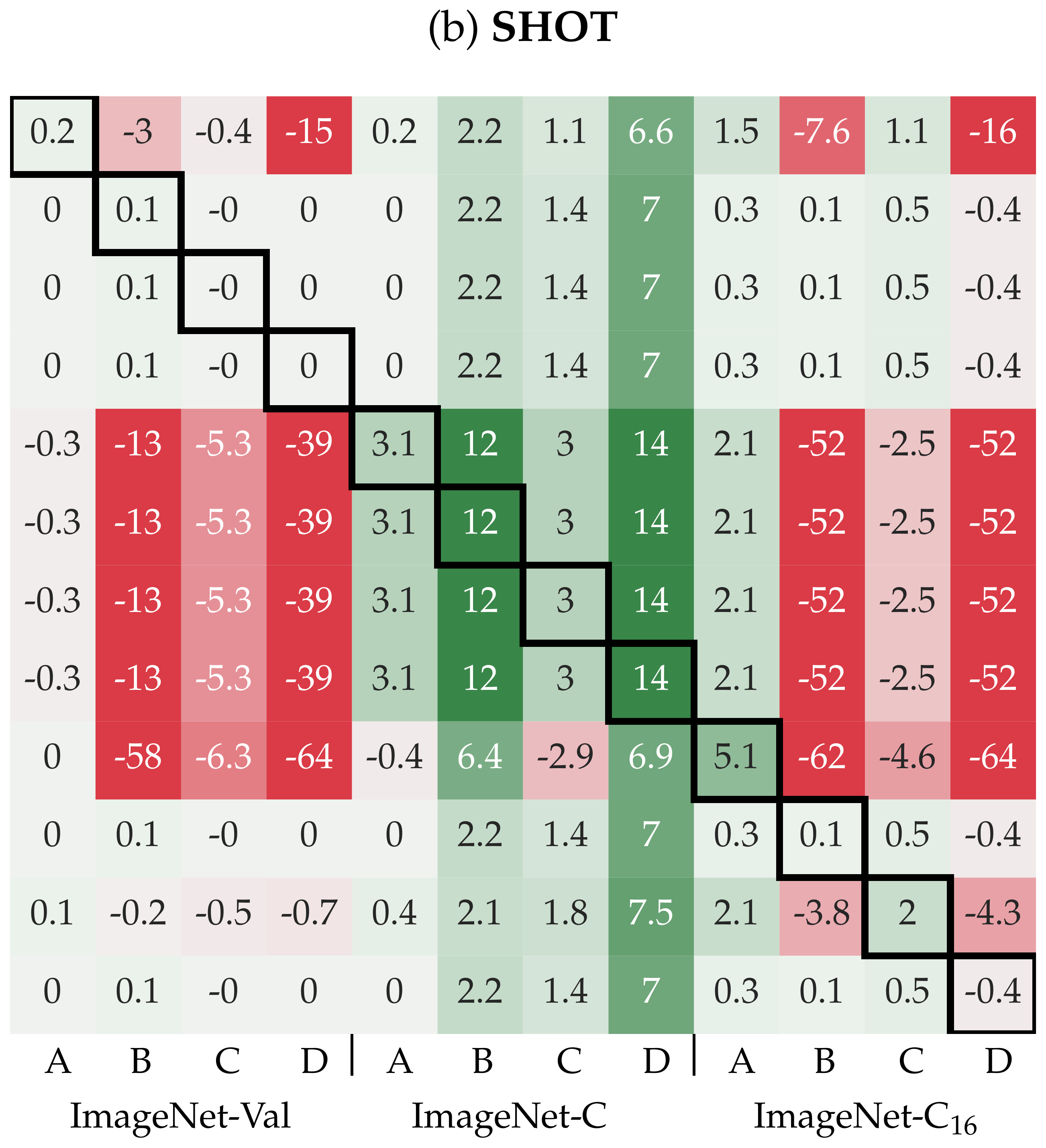} \\
	\vspace{1.2em}
	\includegraphics[width=0.35\textwidth]{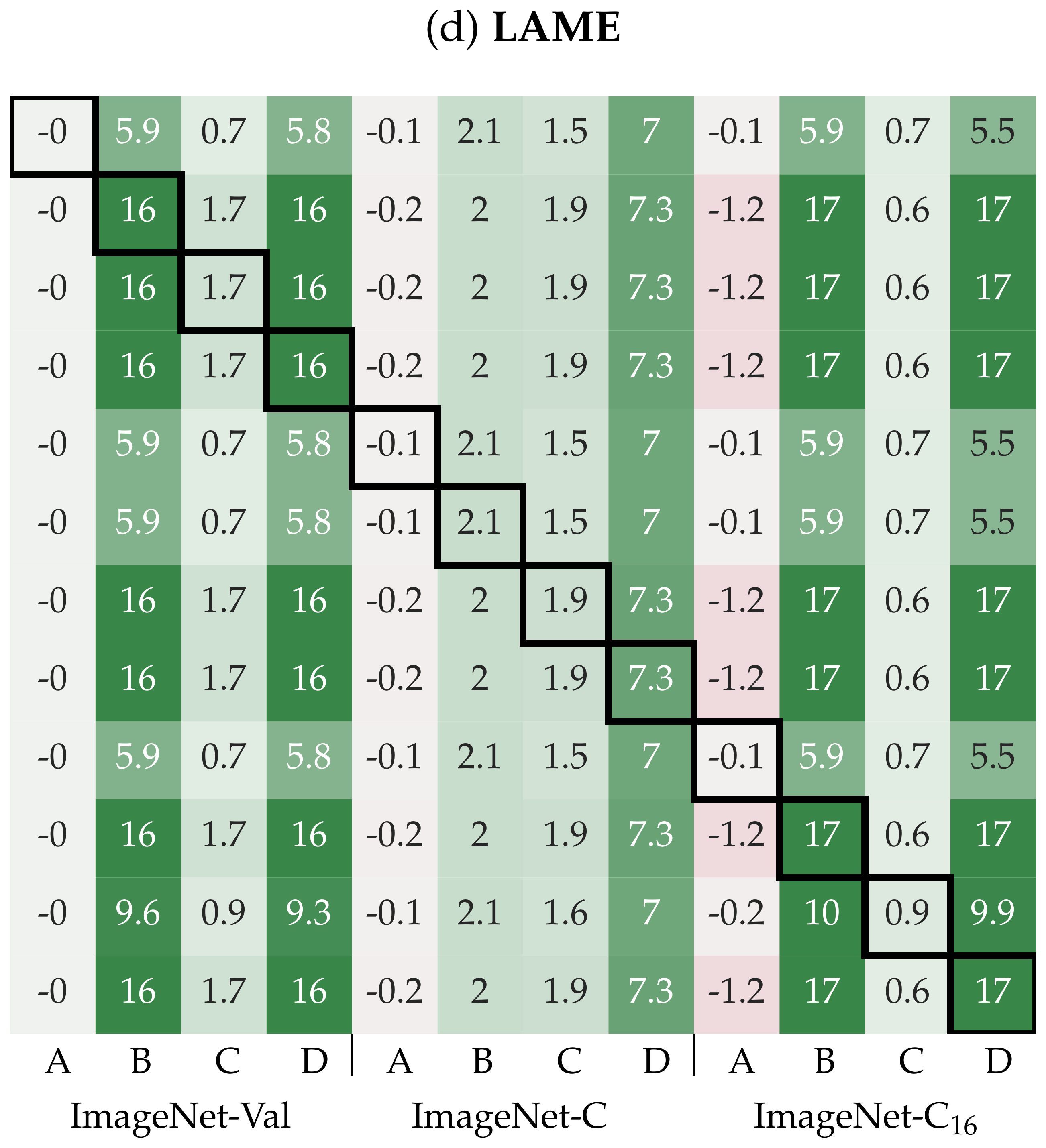} \qquad
	\includegraphics[width=0.35\textwidth]{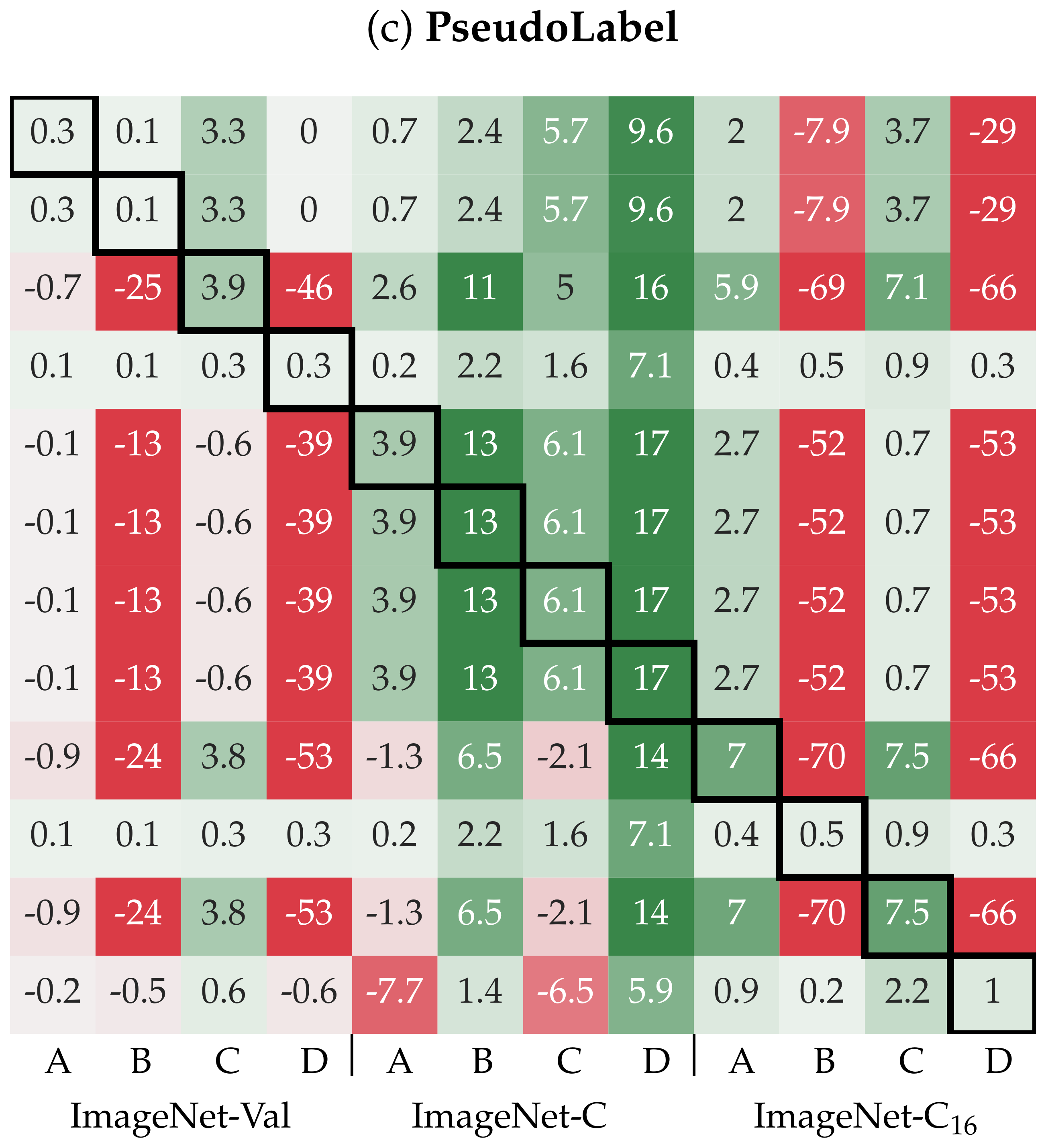} \\
	\vspace{1.2em}
	\includegraphics[width=0.37\textwidth]{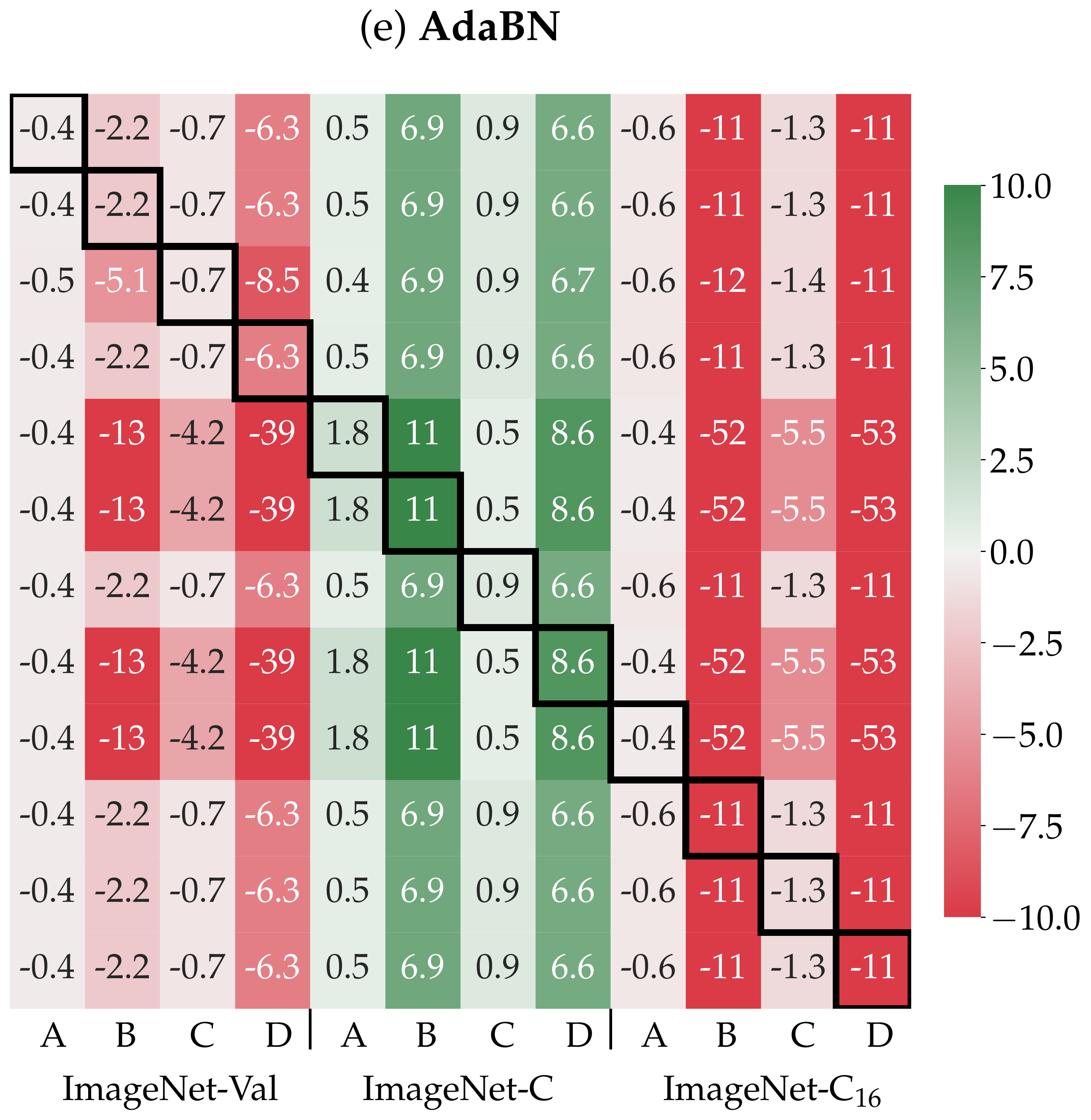}
	\caption{\textit{Cross-shift} validation for all methods. A cell at position $(i, j)$ shows the absolute improvement (or degradation) of the current method \wrt to the baseline when using the optimal hyper-parameters for scenario $i$, but evaluating in scenario $j$. Legend: A = i.i.d.,  B = non i.i.d., C = i.i.d. + prior shift, D = non i.i.d. + prior shift. More details on the scenarios in Sec. \ref{sec:experiments}}
	\label{fig:all_cross_cases}
\end{figure*}

In Fig. \ref{fig:all_cross_cases}, we provide the cross-shift validation matrices for all methods we experimented with. All NAMs suffer from dramatic ``off-diagonal degradation'' when using a single scenario to tune the hyperparameters. This again highlights that, in order to limit the risk of failure at test time, NAMs need to be evaluated across a broad set of validation scenarios. On the other hand, \LAME is much more robust, partly due to the fact that it introduces fewer hyperparameters by design.
One extra hyperparameter that we could tune is a scalar deciding the relative weight of the two terms in the LAME loss of Eq.~(\ref{LAME}).
We expect that, by tuning this extra hyperparameter, we would achieve overall higher performance for the main experiments, but also slightly worse off-diagonal degradation in the confusion matrix.

\section{Ablation study on the batch size}

\begin{figure}[h!]
	\centering
	\includegraphics[scale=0.225]{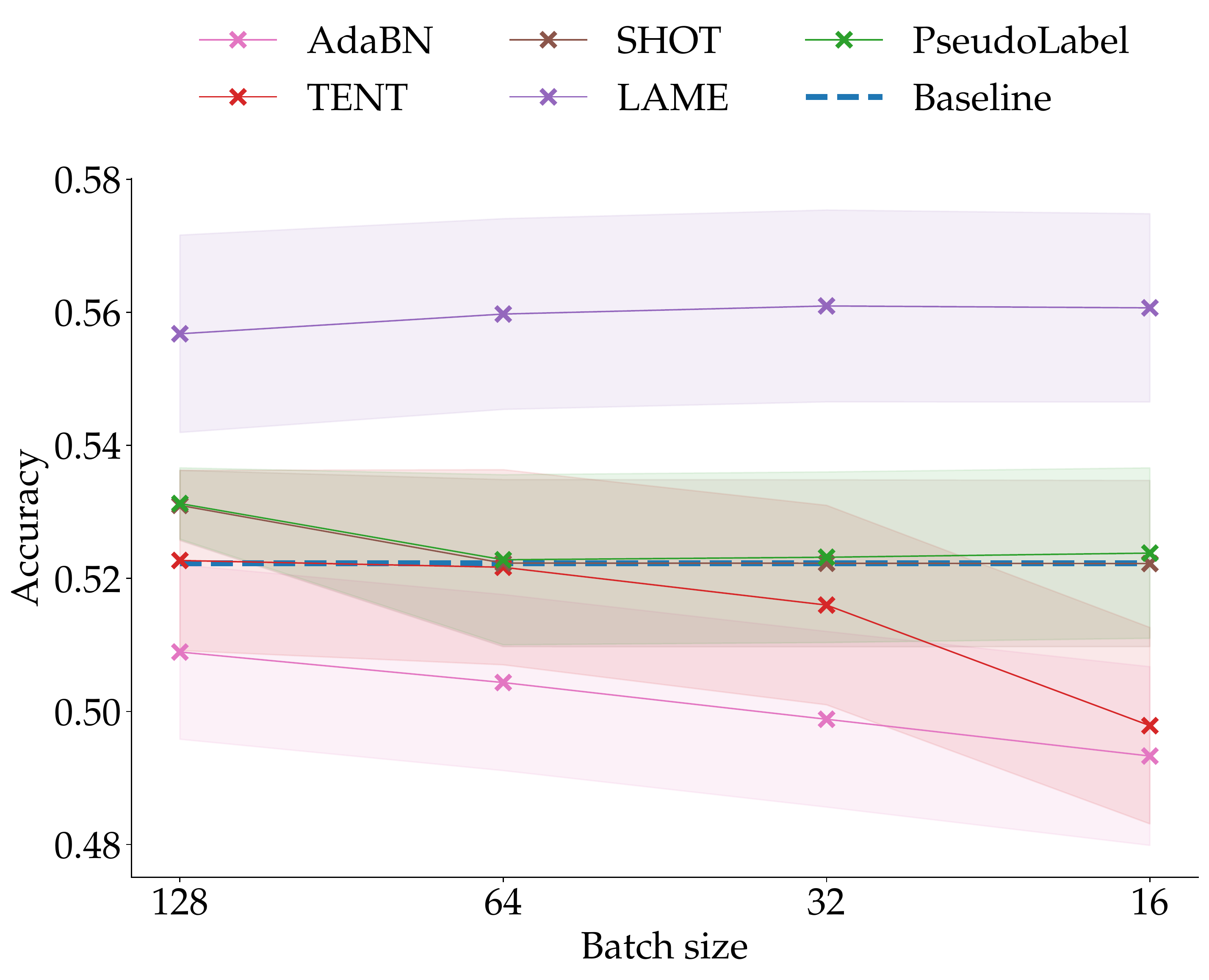}
	\caption{Average accuracy across 7 test scenarios versus batch size, using the Original RN-50. Above 128, NAMs do not fit on a consumer-grade standard 11 GB GPU. In fact, for larger architectures such as ViT-B, a batch size of 16 is already enough to max out memory. Therefore, performing well in the low batch-size regime is a highly desirable property for TTA methods.}
	\label{fig:batch_size}
\end{figure}

Given that \LAME essentially performs output probability correction at the batch level, it is quite important to assess the influence of the size of the batches on the performance of the method. Fig.~\ref{fig:batch_size} confirms that $\LAME$ preserves a close-to-$4\%$ average improvement over the baseline across a wide range of batch sizes.

\section{Datasets}

	In Table \ref{tab:datasets}, we present some characteristics of all the datasets used in our experiments. We group the datasets according to whether they are used for the validation or  testing stage.

\clearpage
\section{Mapping} \label{sec:detail_mapping}

		Our source model produces output probabilities over all the source classes $\mathcal{Y}$, and one needs to map this output to a valid distribution over target classes $\mathcal{Z}$, as motivated in Section \ref{sec:problem_formulation}. Achieving this, as done in Section \ref{sec:formal_link_z_y}, requires the knowledge of a deterministic mapping $\mu$. We hereby describe how we concretely obtain this mapping in our experiments. Recall that we exclusively rely on ImageNet-trained models, such that $\mathcal{Y}$ always corresponds to the set of ImageNet classes. Therefore, we can directly leverage the existing ImageNet hierarchy in order to design $\mu$. Specifically, $\mu$ is defined as follows:

		\begin{align}
			\mu(y) =     \begin{cases}
						      z & \text{if $\exists~z \in \mathcal{Z}: y \in Child(z)$}\\
						      \emptyset & \text{otherwise} \\
					    \end{cases}    
		\end{align}

		\begin{table}
\centering
\caption{Qualitative results obtained with the ancestral mapping scheme for the scenario ImageNet $\rightarrow$ ImageNet-Vid. 279 out of the 1000 original ImageNet were mapped to some superclass.}
\label{tab:AncestralSynsetMapper}
\begin{tabularx}{\columnwidth}{lX}
\toprule
ImageNet-Vid ``superclasses'' & ImageNet classes \\
\midrule
         fox &           kit fox, red fox, grey fox, Arctic fox \\
         dog & English setter, Siberian husky, Australian ter\dots \\
       whale &                         grey whale, killer whale \\
   red panda &                                     lesser panda \\
domestic cat & Egyptian cat, Persian cat, tiger cat, \dots \\
    antelope &                      gazelle, impala, hartebeest \\
    elephant &                African elephant, Indian elephant \\
      monkey & titi, colobus, guenon, squirrel monkey \dots \\
       horse &                                           sorrel \\
    squirrel &                                     fox squirrel \\
        bear & brown bear, ice bear, black bear \dots \\
       tiger &                                            tiger \\
       zebra &                                            zebra \\
       sheep &                                              ram \\
      cattle &                                               ox \\
     hamster &                                          hamster \\
      rabbit &                              Angora, wood rabbit \\
 giant panda &                                      giant panda \\
        lion &                                             lion \\
    airplane &                                         airliner \\
        boat & fireboat, gondola, speedboat, lifeboat \dots \\
     bicycle &             bicycle-built-for-two, mountain bike \\
         car & ambulance, beach wagon, cab, \dots \\
  motorcycle &                                            moped \\
        bird & cock, hen, ostrich, brambling, \dots \\
      turtle & loggerhead, leatherback turtle, mud turtle, \dots \\
      lizard & banded gecko, common iguana, American chameleo \dots \\
       snake & thunder snake, ringneck snake,  \dots \\
         bus &                  trolleybus, minibus, school bus \\
       train &                                     bullet train \\
\bottomrule
\end{tabularx}
\end{table}

		where the function $Child(z)$ outputs the set of all descendants of $z$ in the graph of all ImageNet concepts (or ``synsets''). Note that in the case where a conflict exists, \ie class $y$ has multiple target super-classes as parents, we only keep the closest super-class according to the minimum distance in the graph. We qualitative verify that this ancestral scheme produces a sensible mapping of classes between ImageNet and ImageNet-Vid classes in Table \ref{tab:AncestralSynsetMapper}.

\end{document}